# Multicriteria Optimization Techniques for Understanding the Case Mix Landscape of a Hospital

Robert L Burdett[1,2], r.burdett@qut.edu.au (Corresponding Author)
Paul Corry[1], p.corry@qut.edu.au
Prasad Yarlagadda[3], y.prasad@qut.edu.au
David Cook[4], d.cook@uq.edu.au
Sean Birgan[4], sean.birgan@health.qld.gov.au

[1] School of Mathematical Sciences, Queensland University of Technology, Brisbane, Qld, Australia
[2] School of Mechanical, Medical & Process Engineering, Queensland University of Technology, Brisbane, Qld, Australia
[3] University of Southern Queensland, Springfield, Qld, 4300, Australia
[4] Princess Alexandra Hospital, 2 Ipswich Rd, Woolloongabba, Brisbane, Qld 4102, Australia

**Abstract** – Various medical and surgical units operate in a typical hospital and to treat their patients these units compete for infrastructure like operating rooms (OR) and ward beds. How that competition is regulated affects the capacity and output of a hospital. This article considers the impact of treating different patient case mix (PCM) in a hospital. As each case mix has an economic consequence and a unique profile of hospital resource usage, this consideration is important. To better understand the case mix landscape and to identify those which are optimal from a capacity utilisation perspective, an improved multicriteria optimization (MCO) approach is proposed. As there are many patient types in a typical hospital, the task of generating an archive of non-dominated (i.e., Pareto optimal) case mix is computationally challenging. To generate a better archive, an improved parallelised epsilon constraint method (ECM) is introduced. Our parallel random corrective approach is significantly faster than prior methods and is not restricted to evaluating points on a structured uniform mesh. As such we can generate more solutions. The application of KD-Trees is another new contribution. We use them to perform proximity testing and to store the high dimensional Pareto frontier (PF). For generating, viewing, navigating, and querying an archive, the development of a suitable decision support tool (DST) is proposed and demonstrated.

## 1. Introduction

In this article, we consider the impacts of treating different patient case mixes in a single hospital and building upon past research suggest new mathematical techniques to better quantify that impact. Specifically, we reconsider how to generate Pareto optimal case mix, how to navigate the Pareto frontier (PF) of non-dominated case mix, and how to convey information to hospital managers, planners, and executives so that insights can be more easily understood and actioned. The objective space in a typical hospital capacity assessment (HCA) has high dimension, (e.g., one for each medical and surgical specialty) relative to other decision problems and that intensifies the underlying challenge. Apart from a few exceptions, situations with a higher number of objective functions are seldom addressed. In fact, problems consisting of more than eight objectives are rare and considered challenging, both at the optimization and the human-cognitive levels (Chen et al., 2013).

  Hospital capacity is worth measuring for a variety of pragmatic purposes. We direct readers to Burdett et al. (2017) for the complete details of those. It is worth noting that hospital capacity is regarded as the maximum number or rate of patients that can be treated within a given period. However, realistically there is no single value; it depends on the patient case mix, which can vary in in an unlimited number of ways. For that reason, hospital capacity assessment (HCA) is a challenging task and hospital capacity is difficult to convey.

  Hospital capacity assessment is akin to case-mix planning (CMP) with total patients treated as the objective. Case-mix planning (CMP) is a strategic activity to identify a patient caseload (a.k.a.

cohort) with a specific set of features deemed desirable or ideal. To perform HCA and CMP, it is necessary to partition patients into a distinct set of groups each with a common characteristic. Each group may refer to a particular medical or surgical speciality, a particular patient type, or patients with a particular condition and/or illness. To compute hospital utilization, it is necessary to profile the resource requirements and resource consumption for each group. Both HCA and CMP are multicriteria decision problem because each distinct group of patients has conflicting interests and shares resources (i.e., like operating theatres and in-patient beds) with other groups.

The rest of the paper is organized as follows. In Section 2 the current state of the art is examined, and essential background methodological information is provided. In Section 3 the details of the quantitative framework are provided. In Section 4, a real-life case study is presented. Last, the conclusions, managerial insights and future research directions are detailed.

## 2. Literature Review and Methodological Background

As a foundation for later developments, a brief review of hospital case-mix planning, and capacity assessment is first provided. A detailed review of multicriteria decision making and optimization follows.

### 2.1. Hospital Capacity and Case Mix Planning

Hospital case-mix planning, and capacity assessment are contemporary topics. In recent times there has been considerable interest from researchers, academics, and other decision makers. Deficiencies in existing health care systems and practices, exacerbated by the COVID pandemic, has fuelled research to find better ways to plan and manage health care resources. In past research, a variety of approaches have been applied to the aforesaid decision problem, including mixed integer programming (Ma et al. (2011), Burdett et al. (2017), Shafaei and Mozdgir (2018)), stochastic programming (Neyshabouri and Berg (2017), Freeman et al. (2018), McRae and Brunner (2019), process mining (Andrews at al. (2022)) and multicriteria optimisation (Malik et al. (2015), Burdett and Kozan (2016), Zhou et al. (2018), Chalgham et.al. (2019)).

Table 1 summarises the most crucial details about recent research. In summary, Ma et al. (2011) developed and tested a case mix planning model maximizing the overall financial contribution of a hospital. Malik et.al. (2015) formulated and solved a bi-objective aggregate capacity planning problem for operating theatres. Zhou et al. (2018) considered equity and revenue objectives and the capacity allocation of wards. Burdett et al. (2017) considered an entire hospital and included ward and intensive care bed capacity allocations. Burdett and Kozan (2016) then provided a bona-fide multicriteria approach for HCA. Their grid-based method was able to generate around twenty thousand non-dominated solutions in approximately two days. Similarly, their non-grid based random-corrective method was able to generate ten thousand in about eight hours. Shafaei and Mozdgir (2018) developed a mathematical model to optimize the allocation of OR time among surgical groups and applied a robust estimator for values of the model parameters. McRae, Brunner, Bard (2018) developed a non-linear mixed-integer programming model and incorporated economies of scale. Freeman et al. (2018) developed an iterative approach to generate a set of candidate solutions. They applied simulation techniques to evaluate the master-surgical schedule (MSS) and each case mix solution. McRae and Brunner (2019) presented a framework for evaluating the effect of stochastic parameters on the case mix of a hospital. Chalgham et.al. (2019) proposed multicriteria decision methods to improve in-patient flows from an emergency department. Burdett et al. (2023a) provided the first regional HCA approach and applied it to a 15-hospital regional case study. Saha and Rathore (2022) considered physicians as a significant limiting factor in hospital care. They developed a two-stage stochastic programming approach in which decisions on regular physician allocation and their capacity adjustments are a trade-off between expected cost and patient demand fulfillment. To solve the problem inexactly, a scenario-based heuristic was applied with one thousand scenarios.

**Table 1.** Summary of recent CMP and related research

| Article | Problem | EDD | OR | WARD | ICU | STOCH | MCS | DST | REG | Objective & Method |
|---|---|---|---|---|---|---|---|---|---|---|
| Ma et al. (2011) | CMS | ✗ | ✓ | ✓ | ✗ | ✗ | ✗ | ✗ | ✗ | Profit; MIP; |
| Ma & Demeulemeester (2013) | CMP+ORS | ✗ | ✓ | ✓ | ✗ | ✗ | ✗ | ✗ | ✗ | Profit; Bed Shortage; MIP; |
| Malik et.al. (2015) | ORP | ✗ | ✓ | ✓ | ✓ | ✗ | ✓ | ✗ | ✗ | Waiting List Size; Costs; Meta H.; |
| Jebali and Diabat (2015) | ORP | ✗ | ✓ | ✓ | ✓ | ✓ | ✗ | ✗ | ✗ | Costs; SAA; |
| Burdett & Kozan (2016) | HCA | ✗ | ✓ | ✓ | ✓ | ✗ | ✓ | ✗ | ✗ | Output; LP, ECM; |
| Yahia et al. (2016) | CMP | ✗ | ✓ | ✓ | ✓ | ✓ | ✗ | ✗ | ✗ | Output; SAA; |
| Jebali and Diabat (2017) | ORP | ✗ | ✓ | ✗ | ✓ | ✓ | ✗ | ✗ | ✗ | Cost; SAA; |
| Burdett et al. (2017) | HCA | ✓ | ✓ | ✓ | ✓ | ✗ | ✗ | ✗ | ✗ | Output; LP; |
| Zhou et al. (2018) | HCA | ✗ | ✓ | ✓ | ✗ | ✓ | ✓ | ✗ | ✗ | Revenue; Equity; DES, MIP, ECM; |
| Shafaei & Mozdgir (2018) | ORP | ✗ | ✓ | ✓ | ✓ | ✓ | ✗ | ✗ | ✗ | Value; LP & TOPSIS; |
| Freeman et al. (2018) | ORS | ✗ | ✓ | ✓ | ✓ | ✓ | ✗ | ✗ | ✗ | Payment; MIP; |
| McRae et al. (2018) | CMP | ✗ | ✓ | ✓ | ✓ | ✗ | ✗ | ✗ | ✗ | Profit; NLP; |
| McRae & Brunner (2019) | CMP | ✗ | ✓ | ✓ | ✓ | ✓ | ✗ | ✗ | ✗ | Revenue; SAA; |
| Saha & Rathore (2022) | CMP | ✗ | ✗ | ✗ | ✓ | ✓ | ✗ | ✗ | ✗ | Expected Cost; Heuristic; |
| Burdett et al. (2023a) | HCA | ✗ | ✓ | ✓ | ✓ | ✗ | ✓ | ✗ | ✓ | Output; Unmet Demand; Outsourcing; MIP; |
| Burdett et al. (2023b) | HCA | ✗ | ✓ | ✓ | ✓ | ✗ | ✓ | ✓ | ✗ | Output; MIP; HOPLITE; |
| This article | HCA | ✗ | ✓ | ✓ | ✓ | ✗ | ✓ | ✓ | ✗ | Output; ECM; |

**Key:** CMP: Case Mix Planning; CMS: Case Mix Scheduling; DES: Discrete Event Simulation; DST: Decision Support Tool; ECM: Epsilon Constraint Method; HCA: Hospital Capacity Allocation; LP: Linear Programming; MIP: Mixed Integer Programming; ORS: Operating Room Scheduling; ORP: Operating Room Planning; REG: Regional; SAA: Sample Average Approximation

## 2.2. Multi-Criteria Analysis and Optimization

Multicriteria optimization problems arise in numerous applications. In MCO the identification of the best compromise solutions is most often the predominant focus. It is rarely practical to generate the entire PF, so a subset that best summarises the key features and attributes is instead generated. Another notable step in MCO involves the navigation from one non-dominated solution (NDS) to another and the ultimate selection of a single solution. The field of MCO is mature, and multi-faceted. To supply a comprehensive approach for multicriteria HCA, it is necessary to understand which techniques from the literature are best suited, and whether novel approaches are needed. The main aspects of MCO are now considered in detail.

***Archive Generation.*** The most crucial task in MCO is the identification of Pareto-optimal solutions. These solutions describe trade-offs between the objective functions considered. Generating a PF is computationally intractable in higher dimensions, and it is rarely possible to generate the PF in sufficient detail to supply a comprehensive picture of all inherent trade-offs. As such, a good approximation is sought. Many techniques have been developed in the literature. At present, the literature suggests that there are a set of popular techniques. These are well tested and well used, particularly in engineering, transportation, and economics (Bevrani et al., 2020), though only applied to problems with few objective functions. If the optimization problem is tractable and has few objective functions, then the epsilon-constraint method is well suited. For problems with considerably more functions, adaptive and augmented variants have been created with superior performance. These are applied in Laumanns et al. (2006), Mavrotas (2009), Kirklik and Sayin (2014), Burdett (2015) and Burdett and Kozan (2016). Laumanns et al. (2006) dynamically partition the hyper grid of the objective space using information that is obtained as the ECM method is applied. In so doing, they can more effectively generate NDS and avoid redundant single objective solves. Mavrotas (2009) also avoid redundant solves by exiting from the nested loop of the ECM when the problem becomes infeasible. In Utyuzhnikova et. al (2009) a method for generating a well-distributed Pareto set for both convex and non-convex frontiers was proposed. Their approach is like the Normal-Boundary Intersection method. They remove local Pareto solutions using a simple algorithm. In Kirklik and Sayin (2014), search regions are intelligently tracked, leading to further improvement of the ECM. Ehlers (2015) proposed an algorithm to enumerate the elements of a Pareto front given that the domain is integer valued or can be made so by discretization. The algorithm computes the Pareto solutions successively and is well suited to scenarios where nothing is known about the frontier or its' size.

Klamroth et al. (2015) make use of tight local upper bounds to eliminate redundancies in the ECM. Local upper bounds induce a decomposition of the search region into search zones. Dachert et al., (2017) provide new theoretical insights regarding structural properties of the search regions in MCO. They introduce a neighbourhood structure between local upper bounds. Vakhania and Werner (2022) developed an approach called multi-threshold optimization and applied it to a multi-criteria single machine scheduling problem. The concept of their approach is to provide in polynomial time, a solution with an acceptable quality for a given threshold vector. The threshold vector describes upper and lower bounds for each objective function. Their approach bypasses generation of the PF completely and supplies a pragmatic approach well suited to real decision makers.

For intractable MCO problems and those with non-linearities, meta-heuristics are essential. The Non-Dominated Sorting Genetic Algorithm (NSGA) and the Multi Objective Evolutionary Algorithm (MOEA) are the most noteworthy and prevalent approaches in the literature (Deb and Jain, 2014). Other types of meta-heuristic are also being applied in the literature that exploit types of swarm intelligence. A detailed list of these is present in Khalilporazari et al. (2020). These include whale, firefly, dragonfly, bat, cuckoo, crow, cat, wolf, invasive weed, and sparrow (Li and Wang, 2022). Most articles of that nature, however, supply little contribution to MCO theory. The current research trend is the consideration of MCO problems with complex discontinuous and irregular Pareto fronts, with different degrees of convexity and concavity. Cao et al. (2022) developed a curvature-based PF estimation method for MOEAs. Feng et al. (2021) also developed a MOEA but with a perception property. Zhang et al. (2018) created a penalty-based boundary intersection approach for a MOEA. Elarbi et al. (2020) proposed a decomposition-based algorithm for many-objective optimization problems. They employ predefined normal boundary intersection directions to generate a well distributed PF.

***Archive Storage and Management.*** NDS need to be stored and organized intelligently within an archive. How to do these things is an important research theme. The approaches described in Fieldsend et al. (2003) report that rapid searching of an archive is achievable with logarithmic complexity using binary trees. In one approach they created a binary tree for each dimension. They also introduced dominated and nondominated trees.

From a decision makers perspective, any reduction in the size of the PF is most helpful and inspecting and critiquing fewer potential answers is much simpler. Reducing a PF to a more manageable size is explored in numerous papers. Without loss of generality, most methods remove solutions from dense regions. One set of approaches prunes non-dominated solutions using the concept of hypercubes (Laumanns et al. (2002, 2006)). Using the concept of e-dominance, the "objective space" can be partitioned into uniform homogenous hypercubes (a.k.a., e-boxes) and populated with a chosen number of solutions, often one. Other approaches use clustering and crowding distance (Patil, 2018). Cheikh et al. (2010) proposed a model to select a restricted set of solutions from a set of Pareto optimal solutions. The method clusters solutions into groups (i.e., using meta-heuristics), each with comparable properties. A representative solution is then chosen from each cluster. In their approach, the Euclidean distance is used as a metric of similarity. This means that solutions are considered similar if they are sufficiently close. This is a potential limitation because there exists other similarity measures.

***Archive Updating***. Archive updating is vital in methods that generate NDS. In meta-heuristic approaches, the archive is revised as solutions are created. A variety of methods have been developed to date. Knowles and Corne (2003) introduced archiving strategies to keep an archive of bounded size, with an even distribution of points across the Pareto front. Knowles and Corne described their approach as adaptive and computationally efficient, suitable for use with any Pareto optimization algorithm. At each iteration, the algorithm generates one point and updates the archive as specified by a generic function. Glasmachers (2017) proposed an archiving algorithm based upon *k-d* tree for binary space partitioning. They update an archive of Pareto optimal vectors iteratively in a sequence. Their approach was shown to be effective for medium to large scale archives and intended for online processing of non-dominated sets in evolutionary multi-objective optimization. A hyper grid management scheme for external archive management was evaluated in Patil (2018). The distinguishing feature of their approach is the avoidance of grid boundary recalculations.

***Archive Characterisation.*** Constructing a model of the Pareto frontier has been considered in the literature. Daskilewicz (2013) and Daskilewicz and German (2014) developed an approach to parameterize a sampled PF using a self-ordered map (SOM). Their paper describes a process for defining a barycentric coordinate system that parameterizes the entire sampled PF. Hancock et al. (2015) considered a weakness of the current e-dominance mechanism, which is the inability to vary the resolution it provides of the PF based upon the frontier's trade-off properties. They introduced and tested the concept of L-Dominance. Kobayashi et al. (2018) proposed a "Bezier simplex model" and a fitting algorithm to decompose a high dimensional surface into an accurate approximation with a smaller sample. Their approach was able to fit a four-objective real-world example with 58 points.

***Archive Quality.*** Measuring the quality of an archive relative to the true PF or another archive is an worthwhile task. At present, computing the hypervolume of an archive of NDS is suggested (Zitzler et al., 2007, Cao et al., 2015). The hypervolume measures the size of the space enclosed by all solutions in the archive and a user-defined reference point. However, there are numerous comments that it is impractical for problems with higher dimensions (While et al., 2006).

***Selecting NDS.*** Once an archive of NDS has been obtained, a crucial task is to choose an acceptable solution. Critical to making that decision, however, is the assumption that the Pareto front has been sufficiently populated (Cao et al., 2015). There are various methods that are used to rank the NDS generated. Wang and Rangaiah (2017) and Wang et al. (2020) for instance have quite recently analysed (TOPSIS, LINMAP, VIKOR), (SAW, MEW), (ELECTRE, NFM), and (FUCA, GRA) methods in relation to a chemical engineering decision problem. Various concepts and philosophies are used as the basis of these ranking methods; however, most are subjective, and supply different answers to the same situation.

***Archive Visualization.*** The set of NDS can be vast. Comparing alternatives (i.e., multi-dimensional vectors) and visualizing the trade-offs that occur is problematic, particularly in decision problems with many objectives. Miettinen (2014) has previously demonstrated that there are many visualization tools, including bar charts, scatter plots, value paths (i.e., parallel coordinates plot), multiway dot plots, star coordinate systems, spider web charts, petal diagrams, etc. They found each has benefits and weaknesses, but none are universally better, or worse. The biggest weakness as Daskilewicz and German (2014) reports, is that existing high dimensional visualization techniques do not necessarily offer a clear and natural view.

In two or three objectives, a 2D or 3D scatter plot is the most intuitive way to visualize the results (Talukder and Deb, 2018). Dimensionality reduction is often done for visualisation. For higher dimensions, most approaches in the literature are based upon the mapping of solution points onto a lower dimensional space and include projections and slicing. A variable can be discarded from consideration by orthogonally projecting the data set onto the lower-dimensional subspace that excludes this variable's dimension. A design variable may be fixed at a particular value, reducing the dimensionality of the design problem by one (Daskilewicz, 2013).

A popular approach for visualising Pareto fronts is the application of self-ordered maps. These have topology-preserving properties (Chen et al., 2013). Yoshimi et al. (2015) proposed the application of spherical self-ordered map. They report that "plane" SOMs can cause distortions of data along its edges, whereas "spherical" SOMs, because they have no edges, can visualize similarities among data more accurately. Talukder and Deb (2018) proposed an alternative hierarchical approach. Solutions are clustered and divided into layers according to their centrality within each cluster. They decompose the point cloud into layers of convex hulls and each convex hull layer is plotted on a three-dimensional RadVis plot (with colour), where the position of each layer is defined on the z-axis. They have analysed the pros and cons of previous visualisation methods.

***Archive Interaction and Navigation.*** Interactive methods that require decision makers to participate in the solution process, may be developed for MCO. Hakanen et al. (2021) investigated the application of visual interaction techniques and identified seven high-level tasks of a decision maker (i.e., compare Pareto optimal, specify preferences, check feasibility of preferences, learn about characteristics, detect correlations, post-process most preferred solution). Nine lower-level tasks were identified to facilitate those high-level tasks. They then tested various interactive methods on a wastewater treatment plant example. Laurillau et al. (2018) developed a slider widget called the Trade-Off-Pareto-Slider (i.e., TOP-Slider) to help find optimal compromises between conflicting criteria. An underlying optimization algorithm is hidden from users. The idea behind the use of sliders is that users may explore the solution space through a what-if process. Their widget, however, only considered three objectives. Kangas and Miettinen (2018) considered MCO where some parameters are uncertain. They proposed an interactive method to find a balance between robustness and nominal quality using the concept of light robustness. In their modelling approach a scalarizing function is used to map multiple objectives to a single expression.

***Findings.*** Multi-criteria optimization is an extensively researched and highly contested field. There are few follow up articles, however, which apply existing methodologies. As such, it is difficult to identify the best approach from amongst competitors. Most case studies in recent papers have a small number of objectives but tend to be non-linear. The railway and health applications in Burdett (2015) and Burdett and Kozan (2016) with 24 and 21 objectives respectively, however, are still some of the largest MCO problems considered to date. There are, however, no non-linearities present in those problems. Our literature review indicates that new methods may still be needed. The biggest limitations are still the following:

  i.   Archive Generation. It is unclear how to generate NDS efficiently if the number of objectives is large.
 ii.   Archive Members. It is unclear which NDS should be stored in the archive.
iii.   Archive Navigation. It is unclear how to navigate the archive after one is created.

Data management techniques are essential if the archive of NDS is large. An archive of one hundred thousand Pareto optimal solutions is unlikely to be an issue, however orders of magnitude larger are expected to bring additional complexities (i.e., time to read and write, and memory allocation). The application of ECM and other improved versions which traverse a structured mesh are still inadequate as there are an abundant number of NDS that can be identified. The approaches based upon clustering are limited in the sense that the number of clusters selected is subjective. This affects the solutions that are "singled" out. Clustering is often performed relative to Euclidean distance and that metric is not a "bulletproof" measure of similarity. Solutions must be normalised first, otherwise the Euclidean distance metric can amplify differences and skew results. A more detailed analysis of trade-offs is seldom evaluated.

## 3. Mathematical Approach and Methodology

This section outlines the proposed methodology to perform multicriteria hospital capacity assessment and case-mix sensitivity analysis. After the formal definition of the problem is provided, algorithms and methods for generating Pareto optimal solutions and navigating the PF are introduced.

### 3.1. Formal Problem Definition

Formally we define the decision problem as follows. There is a set of predefined patient type groupings $G$. Within each group $g \in G$ there are subgroups (a.k.a., subtypes). For group $g$ the set of subtypes is designated $P_g$. Each patient subtype treated in the hospital has a set of medical and surgical activities, and each activity requires specific hospital resources for specific amounts of time. The set of activities for patients of type $g$ and subtype $p$ is denoted $A_{g,p}$. For activity $a \in A$ where $A = \bigcup_{g \in G} \bigcup_{p \in P_g} A_{g,p}$, the average treatment time is $t_a$ hours and the set of resources required (i.e., the resourcing profile) is denoted $R_a \subset R$. Set $R_a$ is defined relative to the type of activity being performed. The set of hospital resources $R$ minimally includes the facilities of the hospital like operating rooms, wards and their in-patient beds, and intensive care units and their in-patient beds.

We would like to simultaneously maximize the number (or average throughput) of patients of each type (i.e., $n_g$) treated within a period of $\mathbb{T}$ weeks. The number of patients treated of each subtype is denoted $n_{g,p}$ and $n_g = \sum_{p \in P_g} n_{g,p}$. These decision variables are continuous rates of output over time and do not refer to discrete patients. The total number of patients treated is the output of the hospital and computed as $\mathbb{N} = \sum_{g \in G} n_g$. The output of the hospital is restricted by the resources present, their time availability, and their function. As such, it is necessary to identify a resource allocation, describing which resources will be used to treat each patient. The resource allocation is denoted by $\beta_{a,r}$. This decision variable describes how many patients with activity $a$ are treated by resource $r$. The optimization model for this decision problem is as follows:

Maximize $\tilde{n} = (n_1, n_2, \ldots, n_{|G|})$ (1)

Subject To:

$n_{g,p} = \sum_{r \in R_a} \beta_{a,r}$ $\quad \forall g \in G, \forall p \in P_g, \forall a \in A_{g,p}$ (2)

$\sum_{a \in A_r} \beta_{a,r} t_a \leq T_r$ $\quad \forall r \in R$ where $T_r = h_r \times \mathbb{T}$ (3)

$n_{g,p} \geq \mu_{g,p} n_g$ $\quad \forall g \in G, \forall p \in P_g$ (4)

$n_g, n_{g,p} \geq 0$ $\quad \forall g \in G, \forall p \in P_g$ (5)

$\beta_{a,r} \geq 0$ $\quad \forall a \in A, \forall r \in R_a$ where $\beta_{a,r} = 0$ $\forall a \in A, \forall r \in R \setminus R_a$ (6)

This model is hereby called the "capacity allocation model" (CAM). This model has several essential bookkeeping constraints. Constraint (2) defines the inherent relationship between the number of patients $n_{g,p}$ and the resource allocation. Resource usage is restricted by the time availability of the resource as shown in constraint (3). The time availability weekly of resource $r$ is denoted $h_r$. If the resource is a facility like a ward or intensive care unit, then this number must be multiplied by the number of beds present. The subtype mix is enforced by (4), where proportions $\mu_{g,p}$ are chosen such that $\sum_{p \in P_g} \mu_{g,p} = 1$. Constraints (5) and (6) enforce positive values of the decisions.

The maximum number of patients of each type that can be treated is denoted $\bar{n}_g$. These values can be obtained using the CAM. That involves the repeated solution of the model, considering a distinct group is treated:

$\bar{n}_g = Maximize\ n_g$ s.t. Constraint (2)-(6) and $n_{g'} = 0\ \forall g' \in G | g' \neq g$ (7)

For each group, the lower bound is typically zero, however, there are occasions where trade-offs may not be required. A lexicographic bound analysis (Mavrotas, 2009) can be performed to identify those situations, as follows:

$\underline{n}_g = \min_{g' \in G \setminus \{g\}} (\bar{n}_{g|g'})$ where $\bar{n}_{g|g'} = Maximize\ n_g$ s.t. Constraint (2)-(6) and $n_{g'} \geq \bar{n}_{g'}$ (8)

In (8), $\bar{n}_{g|g'}$ is the maximum achievable output for group $g$ given maximum output for group $g'$. If the group shares resources with other groups, the lower bound will be zero. Otherwise, the lower bound will equal the upper bound.

### 3.2. Generating Non-Dominated Case Mix

In this section efficient methods for generating an archive of non-dominated caseload solutions are pursued. The archive that we seek to generate is a subset of the entire PF as the generation of the whole frontier is impractical in most situations. To generate an archive the repeated solution of the following model is essential:

$$\text{Maximize } n_{g^*} \text{ Subject to: Constraint (2)-(6) and } n_g \geq \varepsilon_g \ \forall g \in G | g \neq g^* \quad (9)$$

Parameter $\varepsilon_g$ represents a minimum requirement for objective $g$. In practice, a hyper-grid with uniformly distributed points is generated and the model is solved for each one of those points. That approach is called the epsilon constraint method (ECM). It is necessary to define a sequence of values for $\varepsilon_g$. A standard approach is to define $\varepsilon_g = \bar{n}_g, \bar{n}_g - \delta_g, \bar{n}_g - 2\delta_g, \dots, \delta_g \ \forall g \in G \setminus \{g^*\}$ where $\delta_g = \bar{n}_g / Z_g$. To obtain a good approximation of the PF, it is necessary to choose $Z_g$ sufficiently large but such an approach is intractable in higher dimensional problems.

The formulation at (9) determines the boundary between feasibility and infeasibility within the objective space and includes both strongly and weakly efficient solutions. Weakly efficient solutions can be improved in one objective, without incurring a deterioration in another, but there is no solution that is strictly better in all objectives (Mavrotas et. al., 2009)). The value of generating weakly efficient solutions is dubious and as such, Mavrotas et. al. (2009) suggested the following approach:

$$\text{Maximize } n_{g^*} + \lambda \sum_{g \in G \setminus \{g^*\}} (n_g - \varepsilon_g) / \bar{n}_g \text{ where } \lambda \cong 0.001 \quad (10)$$

The introduction of the second term ensures that values of $n_g$ larger than $\varepsilon_g$ are returned, if doing so does not affect the value $n_{g^*}$.

The model at (9) or (10) may not solve if the grid point is infeasible. This limitation can be overcome by implementing a "dynamic grid generation approach" like the one in Burdett (2015). Their DYNECM approach expands outwards uniformly from the $n_g = 0 \ \forall g \in G$ case mix solution, until infeasibility is detected. As such, vast areas of infeasible grid points can be bypassed. For the DYNECM the number of solutions is dictated by the expansion strategy encoded and the step size $\delta_g$. Another approach is to correct infeasible grid points as in Burdett and Kozan (2016). In that article, a structured mesh does not need to be generated. Grid points are randomly generated, and a second model is applied to find the nearest feasible grid point from which a new efficient solution can be computed. Hence, every iteration of their random corrective epsilon method (RCECM) is productive. RCECM is also advantageous, as users are provided an explicit option to choose how many solutions they would like to receive.

The DYNECM and RCECM are improved versions of the classical ECM for high dimensional instances, but to date they have been implemented as serial search strategies. Improved parallel implementations are introduced in this article, and their details are now described.

**3.3. Parallel Random Corrective ECM**

The main idea of the PRCECM is to use multiple processors to either more quickly generate a given number of points or to generate more points within a specified time. Some additional advanced features are also included. Before the PRCECM is explained it is necessary to detail the improved RCECM underlying it. Given $data = (G, P_g, \mu_{g,p}, A_{g,p}, A, t_a, R, R_a, A_r, T_r)$ and $dvar = (n_g, n_{g,p}, \beta_{a,r})$, the pseudo-code for the RCECM algorithm is shown below:

**Algorithm 1:** RCECM($I, \mathcal{PF}$)
1. Define $\epsilon \leftarrow (0,0,\dots,|G|)$;
2. $\forall i \in \{1,\dots,I\}$: // Generate $I$ points
    a. $\epsilon_g \leftarrow U(0,\bar{n}_g) \ \forall g \in \{2\dots|G|\}$ ; // Choose grid point randomly
    b. $mod \leftarrow \text{GenerateECMModel}(\epsilon)$;
    c. if($\neg \text{SolveModel}(mod)$) // Grid point was infeasible
        i. $mod \leftarrow \text{GenerateFeasGridPtModel}(\epsilon)$;

        ii. SolveModel($mod$);     // Solve model and extract decision variables
       iii. $\epsilon_g \leftarrow n_g \;\; \forall g \in G$     // Define new epsilon vector
       iv. $mod \leftarrow$ GenerateECMModel($\epsilon$);
       v. SolveModel($mod$);
3. if$\bigl(\neg \mathcal{PF}.\text{is\_in}(n) \land \text{NoCloseNeighbours}(proximity, \mathcal{PF}, n)\bigr)$ $\mathcal{PF}.\text{insert}(n)$;

Parameter $I$ is the intended number of points to generate and store. The predominant step in this algorithm is the evaluation of a randomly chosen grid point defined by $\epsilon$. If the model does not solve, a second model is applied. The two models are constructed as follows:

**Algorithm 2:** GenerateECMModel($data, dvar, \epsilon$)
1. $mod \leftarrow$ new Model($dvar$); // Create model object and initialise decision variables
2. AddCoreConstraints($mod, data, dvar$); // Add capacity allocation constraints (2) - (6)
3. AddEpsilonConstraints($mod, dvar, \epsilon$);
4. $mod.obj \leftarrow$ Maximize $n_1 + \lambda \sum_{g \in \{2 \ldots |G|\}} (\epsilon_g - n_g)/\epsilon_g$;

**Algorithm 3:** AddEpsilonConstraints($mod, dvar, \epsilon$)
1. $\forall g \in \{1, \ldots, |G|\}$: $mod.\text{add\_constraint}(n_g \geq \epsilon_g)$;

**Algorithm 4:** GenerateFeasGridPtModel()
1. $mod \leftarrow$ new Model($dvar$); // Create model object and initialise decision variables
2. AddCoreConstraints($mod, data, dvar$); // Add capacity allocation constraints (2) - (6)
3. $\forall g \in \{1, \ldots, |G|\}$: $mod.\text{add\_constraint}(n_g \leq \epsilon_g)$;
4. $mod.obj \leftarrow$ Minimize $\sum_{g \in \{2, \ldots, |G|\}} (\epsilon_g - n_g)$;

It is important to note that $\mathcal{PF}$ is an object and a reference to the tuple $(\mathcal{T}, \mathcal{A})$ where $\mathcal{A} \subset \mathbb{R}^{|G|}$ (i.e., a simple list of points) and $\mathcal{T} = \text{KDT}(\mathcal{A})$, where $|\mathcal{T}| = |\mathcal{A}|$. KD-Tree (KDT) data structures are binary trees (i.e., each node can have a left and right child node), designed for storing, managing, and querying an archive of points. More details about KD-Trees will be presented later. Once an efficient point is identified, it is recorded or rejected. The "insert" function adds the point to the archive $\mathcal{A}$ (i.e., $\mathcal{A} \leftarrow \mathcal{A} \cup \{n\}$) and to the KDT $\mathcal{T}$. The NoCloseNeighbours procedure identifies if any points are within a specified proximity. This restriction has been introduced to ensure a more uniformly distributed Pareto frontier is obtained. This modification provides the user with a mechanism to restrict points from occurring within an area of the objective space. The proximity parameter is selectable, and smaller values permit a more finely grained frontier to be generated. The use of larger values may permit an archive of a more manageable size to be generated. The pseudo-code is follows:

**Algorithm 5:** bool $\leftarrow$ NoCloseNeighbours($proximity, \mathcal{PF}, n_g$)
1. $\mathcal{PF}.\text{get\_neighbours}(proximity, n_g, found)$;
2. return $(|found| = 0)$;

A major limitation of the RCECM is that only one solution is generated at a time. Using parallel processing many more solutions can be generated, and the Pareto frontier can be approximated quicker. Two improved versions have been investigated. These rely upon the application of the afore-described RCECM. The first variant is as follows:

**Algorithm 6:** PRCECM01($I, J, S$)
1. Define $J$ threads and create Pareto frontiers objects $\mathcal{PF}_j = (\mathcal{A}_j, \mathcal{T}_j) \;\; \forall j \in \{0, 1, \ldots, J\}$;
2. $max\_stage \leftarrow (I/J)/S$;
3. $terminate \leftarrow false$;
4. $step \leftarrow 1$;
5. $while(\neg terminate)$:

a. $\forall j \in \{1, \ldots, J\}$: // Run $J$ different generators in parallel
      i. $\mathcal{PF}_j.\text{clear}(\ )$; // Empty $\mathcal{A}_j$ and $\mathcal{T}_j$
      ii. $\text{RCECM}(S, \mathcal{PF}_j)$; // Generate points using thread $j$
   b. $\forall j \in \{1, \ldots, J\}$: // Standard sequential loop. Consolidate archives
      i. $\forall k \in \{1 \ldots |\mathcal{PF}_j|\}$: // Merge points from the jth frontier
         1. $n \leftarrow \mathcal{PF}_j[k]$; // kth point in archive $\mathcal{PF}_j$
         2. if$(\neg \mathcal{PF}_0.\text{is\_in}(n) \wedge \text{NoCloseNeighbours}(proximity, \mathcal{PF}_0, n))$
            $\mathcal{PF}_0.\text{insert}(n)$;
   c. $stage \leftarrow stage + 1$;
   d. $terminate \leftarrow (stage \geq max\_stage)$ ;

This algorithm generates and evaluates $I$ points in total, and this is done using $J$ independent generators (a.k.a. threads). To each generator an independent archive is established and stored as a KD-Tree. At each step (a.k.a., stage), each thread does a limited (a.k.a., reduced) search and generates a specified number of points $S$ (a.k.a., #PTS_THRD_STAGE), some of which are pruned, due to proximity restrictions. After that activity, all sub-frontiers are aggregated and stored within the master archive, namely $\mathcal{PF}_0$. The process is repeated until a specified number of stages have been completed. At the start of each stage, each generator empties its archive and starts again. The central archive could in theory be re-constructed and balanced for an extra speedup. The KD-Tree of each thread does not need to be rebalanced, because it is emptied regularly. However, if the number of points generated within each stage is too large it may be necessary to reconsider how often rebalancing should be performed. The main limitation of this algorithm is that the proximity check is not performed in parallel; but this course of action, permits comprehensive proximity checking to be accomplished. A nimbler variant where this is relaxed (i.e., proximity testing is not comprehensive) is as follows:

**Algorithm 7 :** PRCECM02$(I, J, S)$
1. Define $J$ threads and create Pareto frontiers $\mathcal{PF}_j \ \forall j \in \{0, 1, \ldots, J\}$;
2. $max\_stage \leftarrow (I/J)/S$;
3. $terminate \leftarrow false$;
4. $stage \leftarrow 1$;
5. $while(\neg terminate)$:
   a. $\forall j \in \{1, \ldots, J\}$: // Parallel for-loop
      i. $\mathcal{PF}_j.\text{clear}(\ )$; // Empty $\mathcal{A}_j$ and $\mathcal{T}_j$
      ii. $\text{RCECM}(S, \mathcal{PF}_j)$; // Generate points using thread $j$
      iii. $\forall k \in \{1 \ldots |\mathcal{PF}_j|\}$:
         1. $n \leftarrow \mathcal{PF}_j[k]$; // kth point in archive $\mathcal{PF}_j$
         2. if$(\mathcal{PF}_0.\text{is\_in}(n) \vee \neg \text{NoCloseNeighbours}(proximity, \mathcal{PF}_0, n))$
            $\mathcal{PF}_j.\text{remove}(n)$;
   b. $\mathcal{PF}_0 = \text{KDT}(\bigcup_{j \in \{0, \ldots, J\}} \mathcal{A}_j)$; // Consolidate archives and build new KD-Tree
   c. $stage \leftarrow stage + 1$;
   d. $terminate \leftarrow (stage \geq max\_stage)$ ;

**Final Remarks**. When using the RCECM and PRCECM, the number of points to select, namely $I$, is highly influential in the quality of the Pareto frontier that is generated. The availability of more threads is always beneficial. The proximity parameter is very influential. If the value is too large, the desired number of points may not be attainable. In other words, the Pareto frontier may be describable using less points. If the value is small, then the requested number of points may not be big enough to adequately describe the Pareto frontier.

## 3.5. Archive Management and Querying

In this section an approach to manage, query, navigate and visualize a large archive of non-dominated caseloads is proposed and several computer implementations are detailed. To explain the approach the following terminology is introduced:

$K$: The number of objectives.
$a$: A solution point, where $a \in \mathbb{R}^K$ and $a = (a_k | a_k \in \mathbb{R})_{k \in \{1,\ldots,K\}}$.
$a_k$: The $kth$ value of a point
$a^{\mathrm{r}}$: A reference point
$a^{\mathrm{I}}, a^{\mathrm{N}}$: The ideal (a.k.a., utopia) and least-ideal (a.k.a., nadir) solution points.
$\mathcal{A}$: A list of solutions; $\mathcal{A} = (a[1], a[2], \ldots, a[|\mathcal{A}|]) \equiv (a[n])_{n \in \{1\ldots|\mathcal{A}|\}}$. The size is $|\mathcal{A}|$.
$\mathcal{PF}$: A list of Pareto optimal solutions.
$\mathcal{C}$: A list of candidate solutions, where $\mathcal{C} \subset \mathcal{PF}$.
$\mathcal{H}$: A hypercube, and $\mathcal{H} = \left((lb_k, ub_k)\right)_{k \in \{1,\ldots,K\}}$ where $lb_k \leq ub_k \; \forall k$.
$\mathcal{H}^{\mathcal{PF}}$: A hypercube describing the range of the Pareto optimal solutions in the archive.
$\mathcal{H}^{\mathrm{r}}, \mathcal{H}^{\mathrm{a}}$: A hypercube describing requested and achievable ranges, respectively.

There are various capacity related questions, and the Pareto frontier of non-dominated case mix solutions can be used to answer many. The following procedures have been encoded as part of a general archiving management strategy.

**Table 2.** Procedures in proposed decision support framework

| Procedure | Description |
|---|---|
| Archive :: RangeQuery($\mathcal{H}^{\mathrm{r}}$) $\mapsto$ ($\mathcal{C}$) | Identify existence of solutions with specific characteristics |
| Archive :: RangeQueryExt($\mathcal{H}^{\mathrm{r}}$) $\mapsto$ ($\mathcal{C}, a^*, \mathcal{H}^{\mathrm{a}}$) | Range query with extended outputs |
| Archive :: ComputeBounds ( ) $\mapsto$ ($\mathcal{H}$) | Determine the bounding hypercube of an archive |
| Archive :: Filter ($\mathcal{H}$) $\mapsto$ ($\mathcal{C}$) | Identify solutions within the designated hypercube |
| Archive :: IsIn ($a$) $\mapsto$ ($bool$) | Check if solution $a$ is in the archive |
| Archive :: InRange($\mathcal{H}, a$) $\mapsto$ ($bool$) | Check if solution point $a$ lies in the hypercube |
| Archive :: GetClosest($a^{\mathrm{r}}$) $\mapsto$ ($a$) | Identify solution closest to the reference solution |
| Archive :: FindMin($k$) $\mapsto$ ($z \in \mathbb{R}$) | Return the smallest value in objective $k$ |
| Archive :: FindMax($k$) $\mapsto$ ($z \in \mathbb{R}$) | Return the largest value in objective $k$ |
| Archive :: IsDominated($a$) $\mapsto$ ($bool$) | Check if a particular solution is dominated |
| Archive :: FindNonDominated() $\mapsto$ ($\mathcal{C}$) | Identify all non-dominated solutions |
| Archive :: GetNeighbours($a, radius$) $\mapsto$ ($\mathcal{C}$) | Identify nearby solutions within similar quality |
| Archive :: CheckOptimality($a$) $\mapsto$ ($bool, \mathcal{C}$) | Identify status of solution point and report alternatives |
| Archive :: AnalyseUniformity($\mathcal{A}$) $\mapsto$ ($gap, \mu, \sigma$) | Analyse gaps in each dimension and report gap statistics |
| Archive :: AnalyseSpread($\mathcal{A}$) $\mapsto$ ($\mu, q^1, q^2, q^3$) | Analyse spread and skewness and report quartiles |

The procedures InRange, GetClosest, GetNeighbours and IsDominated take as input a reference point, which may / may not be present in the archive.
     Two archive implementations were developed and contrasted. The first is based upon a list data structure, and the second a KD Tree (KDT). The implementations for each are respectively provided in Appendix A and B and key features are described below.

***List Based Implementation***. The archive was first recorded as a simple unsorted list, i.e., $Archive = (\mathcal{A})$ where $\mathcal{A} = (a[1], a[2], \ldots)$ and $\mathcal{A} \subset \mathbb{R}^K$. The storage size of $\mathcal{A}$ is $K|\mathcal{A}|$. As described, procedure ComputeBounds implies iteration over the cartesian product $\{1, \ldots, K\} \times \mathcal{A}$ via a nested for-loop. Hence it has complexity $O(K|\mathcal{A}|)$. In most applications $K$ is small relative to the magnitude of set $\mathcal{A}$, namely $|\mathcal{A}|$. In this paper $K = 21$. The inner loop should be set up to iterate over elements of $\mathcal{A}$, as this allows us to evaluate each dimension in parallel. If the solutions were stored in a binary heap, then the minimum/maximum could be found in $O(1)$. Building a heap for each objective, however, has $O(K|\mathcal{A}|)$ complexity. Procedure Filter has $|\mathcal{A}|$ steps. However, processing in parallel is possible. Procedure InRange implies $O(K)$ steps at worst, and $O(1)$ at best. Hence, procedure Filter has complexity $O(K|\mathcal{A}|)$. The GetNearstNeighbour procedure is an $O(|\mathcal{A}|)$ process. The computation of distance, however, can be performed in parallel first. If the reference point is the utopia point, then distance computations can be computed (i.e., in parallel) when the archive is input.

If multiple queries are performed, then the filtering needs to be performed repeatedly from scratch. There are no computational savings as the frontier is not stored in a clever way.

***KDT Implementation***: In our implementation the archive is supplemented by a balanced KDT object, hereby denoted $\mathcal{T}$ such that $\mathcal{T} = \text{make}(\mathcal{A})$, $|\mathcal{T}| = |\mathcal{A}|$ and $Archive = (\mathcal{T}, \mathcal{A})$. As $\mathcal{T}$ is a balanced binary tree, it will have height of $\lceil \log_2(|\mathcal{A}| + 1) \rceil$ (or equivalently $\lfloor \log_2(|\mathcal{A}|) + 1 \rfloor$). Every node contains a k-dimensional point and a left and right pointer that provide links to other tree nodes. At each level of the KDT, a different dimension is used as a "splitting hyperplane, permitting efficient partitioning of the space. A KDT object supports the following activities:

**Table 3.** Primary KDT procedures

| Procedure | Description |
|---|---|
| $\text{KDT} :: \text{make}(\mathcal{A})$ | Construct a balanced KDT |
| $\text{KDT} :: \text{insert}(a)$ and $\text{KDT} :: \text{delete}(a)$ | Insert and delete a point |
| $\text{KDT} :: \text{range\_query}(\mathcal{H}) \mapsto (\mathcal{C})$ | Identify points within the specified hypercube |
| $\text{KDT} :: \text{get\_neighbours}(a, radius) \mapsto (\mathcal{C})$ | Identify all points within a prescribed radius |
| $\text{KDT} :: \text{get\_nearest\_neigbour}(a) \mapsto (a')$ | Identify the closest point |
| $\text{KDT} :: \text{is\_in}(a) \mapsto (bool)$ | Check if the point is present |
| $\text{KDT} :: \text{find\_min}(k) \mapsto (z \in \mathbb{R})$ | Identify the minimum value in the given dimension |
| $\text{KDT} :: \text{find\_max}(k) \mapsto (z \in \mathbb{R})$ | Identify the maximum value in the given dimension |
| $\text{KDT} :: \text{is\_dominated}(a) \mapsto (bool)$ | Identify if any points dominate |
| $\text{KDT} :: \text{find\_non\_dominated}(\mathcal{A}) \mapsto \mathcal{PF}$ | Identify all non-dominated points |

Given $N$ points, construction of a balanced KDT is theoretically $O(K \cdot \log_2 N)$ in the worst case if each dimension is pre-sorted (Brown (2015)). If pre-sorting is not performed then complexity is regarded to be $O(N(\log_2 N)^2)$ (i.e., a sort is required at each of the $O(\log_2 N)$ levels). When searching for a solutions existence within the tree, only one solution is inspected per level, and hence $O(\log_2 N)$ steps are required. If the tree is a chain, then all nodes may need to be checked, resulting in worst case complexity $O(N)$. Inserting and deleting has the same complexity for the same reason. When performing a range search the complexity at worst $O(K \cdot N^{1-1/K})$ on average (Lee and Wong, 1977). The nearest neighbour can be found in $O(\log_2 N)$. Given the information above, the complexity of the archive procedures is contrasted below:

**Table 4.** Complexity of KDT procedures. (*) means no general result has been published.

| **Procedure** | **List::** | **KDT::** |
|---|---|---|
| IsIn | $O(|\mathcal{A}|)$ | $O(\log_2|\mathcal{A}|)$ |
| FindMin & FindMax | $O(|\mathcal{A}|)$ | ⋔ (*) |
| ComputeBounds | $O(K|\mathcal{A}|)$ | $O(K\text{⋔})$. |
| RangeQuery | $O(|\mathcal{A}|)$ | $O(K \cdot |\mathcal{A}|^{1-1/K})$ |
| GetNearestNeighbour | $O(|\mathcal{A}|)$ | $O(\log_2|\mathcal{A}|)$ (avg.) |
| GetNeighbours | $O(|\mathcal{A}|)$ | (*) |
| IsDominated | $O(|\mathcal{A}|)$ | $\psi$ (*) |
| FindNonDominated | $O(|\mathcal{A}|^2)$ (worst) | $O(|\mathcal{A}|\psi)$ |

The simplest List::FindNonDominated algorithm uses basic pairwise comparisons and has complexity $O(|\mathcal{A}|^2)$. The more advanced Mishra-Harit (2010) approach has a reported complexity of $O(|\mathcal{A}| \log_2|\mathcal{A}|$ at best and $O(|\mathcal{A}|^2)$ at worst. The complexity of KDT::FindMin, KDT::FindMax, KDT::GetNeighbours and KDT::IsDominated are hard to analyse, however numerical testing indicates significantly better than $O(|\mathcal{A}|)$ performance in practice and probable logarithmic performance. The KDT::IsDominated procedure involves a range search, however, termination occurs immediately after any dominating point is found, so a complete range query is unrequired. Hence, complexity is far better than $O(K.|\mathcal{A}|^{1-1/K})$. To identify dominated points, the KDT algorithm uses the principles of dominance regions. The KDT::GetNearestNeighbour finds a point closest to a prescribed target, not present within the KDT. If the target point is present, then the traditional implementation is inadequate and will only return the target. That is not what we want. In our application it is necessary to analyse in turn, each point within the KDT as a target. We could remove existing points from the KDT to facilitate a call to the traditional algorithm, but that is inefficient, as we would have to re-add them again. Instead, we have chosen to revise the traditional algorithm, with an additional exclusion mechanism that bypasses any node equal to the target.

***Analysing Uniformity and Dispersion.*** The uniformity and dispersion present in archives $\mathcal{PF}$ and $\mathcal{C}$ is worth analysing. How dispersed or evenly spread the solutions are across each objective is also worth noting. Gaps of unusual size in $\mathcal{PF}$ may occur because the archive is incomplete, and the frontier was not generated well. Other gaps in $\mathcal{PF}$ or $\mathcal{C}$ may result, however, because the objective space is non-convex. To obtain gap information, the scores in each dimension should be sorted from smallest to largest. Then, gaps can be measured, and the mean and standard deviation of the gaps computed:

$$gap_{k,n} = a_k[n+1] - a_k[n] \quad \forall k \in \{1,2,\ldots,K\}, \forall n \in \{1,\ldots,|\mathcal{A}|-1\} \tag{11}$$

$$\mu_k = \left(\frac{1}{|\mathcal{A}|-1}\right)\sum_{\forall n \in \{1,\ldots,|\mathcal{A}|-1\}}(gap_{k,n}) \text{ and } \sigma_k = \sqrt{\left(\frac{1}{|\mathcal{A}|-1}\right)\sum_{\forall n \in \{1,\ldots,|\mathcal{A}|-1\}}(gap_{k,n}-\mu_k)^2} \tag{12}$$

Equation (11) and (12) are computationally intensive for large archives with many objectives. Computing gaps is performed by our AnaylseUniformity procedure. The complexity is $O(K|\mathcal{A}|\log|\mathcal{A}|)$ because the sort takes $|\mathcal{A}|\log|\mathcal{A}|$ time and must be performed $K$ times. However, each dimension can be considered independently, thus permitting parallel computations. The mean and standard deviation are not that descriptive unless values are close to zero. A better measure is the coefficient of variation, namely $\sigma_k/\mu_k$, which is a relative measure of the dispersion. Values close to zero imply minor variation. Values greater than one highlight significant deviations from the average gap size.

***Analysing Range Queries.*** It is important to identify a subset of the Pareto optimal solutions (i.e., denoted $\mathcal{H}^a$) that meets the expectations of decision makers using a range query. The expectations specified by $\mathcal{H}^r$ are specific ranges of each objective. Taken together they constitute a hypercube. The requested hypercube $\mathcal{H}^r$ must be positioned within the objective space described by hypercube $\mathcal{H}^{\mathcal{PF}}$. In other words: $lb_k^{\mathcal{PF}} \leq lb_k^r \leq ub_k^r \leq ub_k^{\mathcal{PF}} \quad \forall k \in \{1,\ldots,K\}$. User defined expectations may vary significantly. For those that are too high, no solutions may be identifiable. In that situation, the expectations of a decision maker may need to be relaxed, and a new query initiated. If there are many alternatives, then expectations should be increased, or vice versa if few exist. Figure 1a shows a two-dimensional example and Figure 1b details the process of querying.

The percentage of the frontier that is achievable, computed as $100 \times |\mathcal{C}|/|\mathcal{PF}|$), is worth reporting after procedure RangeQuery is complete. The three hypercubes $(\mathcal{H}^{\mathcal{PF}}, \mathcal{H}^r, \mathcal{H}^a)$ constitute the main pieces of information in our visualisation. They also facilitate a graphical user interface consisting of sliders. The progress level of the best solution $a^*$ is also worth reporting. Any solution from the archive has a progress level computed as follows: $progress = 100(\gamma - \delta)/\gamma$ where $\gamma = \|a^I - a^N\|$ and $\delta = \|a^I - a\|$. Variable $\gamma$ describes the distance between the ideal and least-ideal solutions, whereas $\delta$ is the proximity to the utopia solution. The progress is least when $\delta$ is large. The distance metric is selectable, but the two-norm is appropriate. Normalized solutions should however be evaluated, to avoid scaling issues.

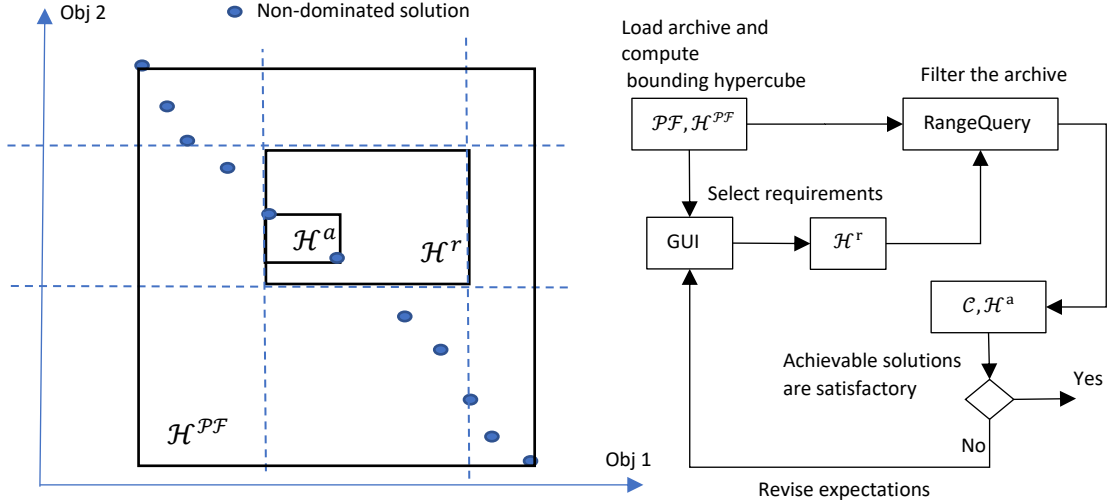

**Figure 1.** Range querying demonstration (i.e., in 2D) and process

The information provided by the three hypercubes (i.e., $\mathcal{H}^p, \mathcal{H}^r, \mathcal{H}^a$) is valuable to decision makers. It can be presented visually using Box and Whisker charts and/or coupled with a textual output. For demonstrative purposes Box and Whisker charts for a 30-point | 3-objective example (see Appendix D), with $\mathcal{H}^{\mathcal{PF}}$ ={[9, 100], [5, 95], [1, 96]}, $\mathcal{H}^r = \{[45,100], [20, 95], [56, 96]\}$ , $\mathcal{H}^a$ ={[68, 100], [26, 93], [76, 96]} are shown in Figure 2. The textual output for that example is as follows:

| | |
|---|---|
| [9, [45, [68, 100]]] | // Objective 1: (9, 45, 68, 100, 100, 100) |
| [5, [20, [26, 93], 95]] | // Objective 2: (5, 20, 26, 93, 95, 95) |
| [1, [56, [76, 96]]] | // Objective 3: (1, 56, 76, 96, 96, 96) |

In this textual format, it is necessary to remove multiple copies of the same value. Also, the bracketing shows how the different intervals are positioned. The inherent ordering, $lb_k^{\mathcal{PF}} \leq lb_k^r \leq lb_k^a \leq ub_k^a \leq ub_k^r \leq ub_k^{\mathcal{PF}}$ $\forall k \in \{1, \dots, K\}$ is also maintained.

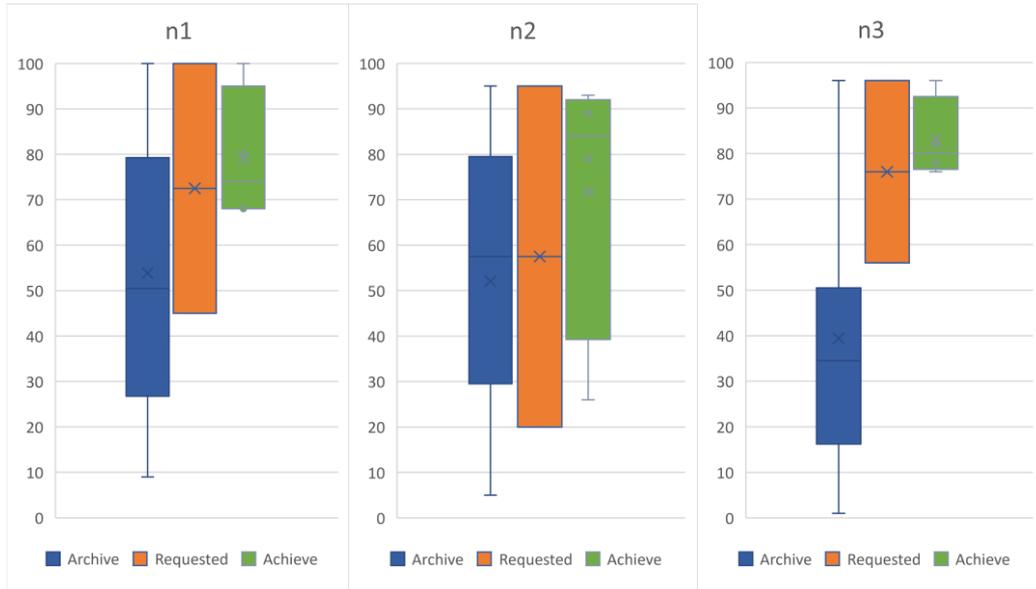

**Figure 2.** Detailed 30-point example showing dispersion.

In this 30-point | 3-objective example, there is no dispersion in the requested range, hence the middle boxplot has a balanced spread. To obtain the charts shown in Figure 2 it is necessary to calculate the

mean and median, and the first and third quartile. For a large archive, with hundreds of thousands, if not millions of points, this can be time consuming. Software like Excel for instance struggles with such a task and is not recommended. A programming language utilising parallel processing should first be used to pre-process the archives. Other visualisation software should only be used for plotting.

To further observe the distribution of the data, violin charts, which are hybrid boxplot and density plots, are particularly well suited. Figure 3 demonstrates their application to the results shown in Figure 2. The width corresponds to the approximate frequency of data points in each region. These charts were plotted using the python "seaborn: statistical data visualization" library (https://seaborn.pydata.org).

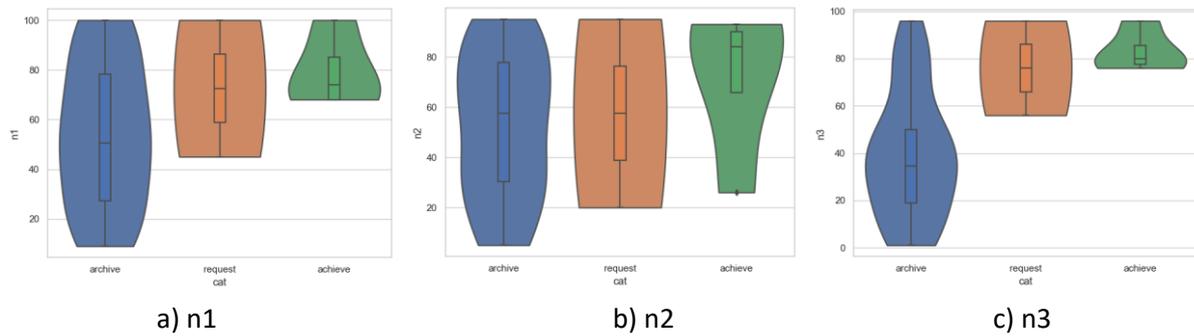

a) n1  b) n2  c) n3

**Figure 3.** Violin plots for 30-point example

Further violin plots of the form shown in Figure 4 are also helpful. Each chart summarises one of the three archives.

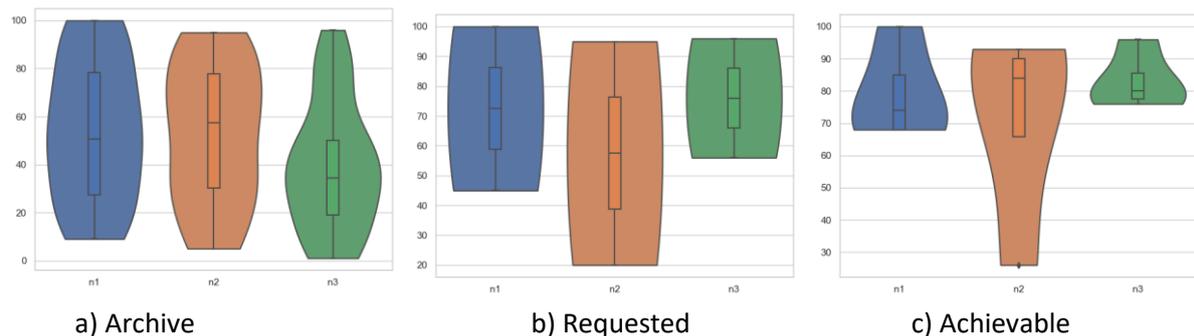

a) Archive  b) Requested  c) Achievable

**Figure 4.** Violin plots of 30-point 3-objective example

***Analysing Case Mix Queries***. A decision maker may suggest a cohort and query whether it is achievable or not. This question may be answered using the archive CheckOptimality and GetClosest procedures. The CheckOptimality procedure reports the existence of better solutions if they exist, or more achievable solutions if the solution point is unachievable. The first step is to call the IsDominated procedure that uses the notion of dominance regions. Depending on the result, one of two range queries is performed. Figure 5 below demonstrates the two situations that may occur for a two-dimensional situation.

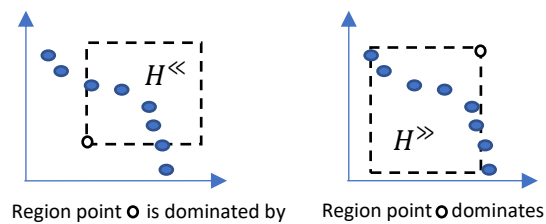

Region point o is dominated by    Region point o dominates

**Figure 5.** Dominance regions in two dimensions assuming max objectives

The point is feasible in the first, but infeasible in the second chart. Assuming maximization objectives, the region point $a$ dominates is $\mathcal{H}_a^{\gg} = \{(lb_k, a_k)\}_{k \in \{1,2,...,K\}}$ and the region that dominates point $a$ is $\mathcal{H}_a^{\ll} = \{(a_k, ub_k)\}_{k \in \{1,2,...,K\}}$ where $\mathcal{H}_a^{\gg}$ and $\mathcal{H}_a^{\ll}$ are $K$ dimensional hypercubes.

### 3.6. Development of a Decision Support Tool

To apply this article's methods, the development of a DST, for health care managers and other planners is considered here. In consideration of that development, a practical workflow is shown in Figure 6. To facilitate that workflow the interfaces in Figure 7 are necessary.

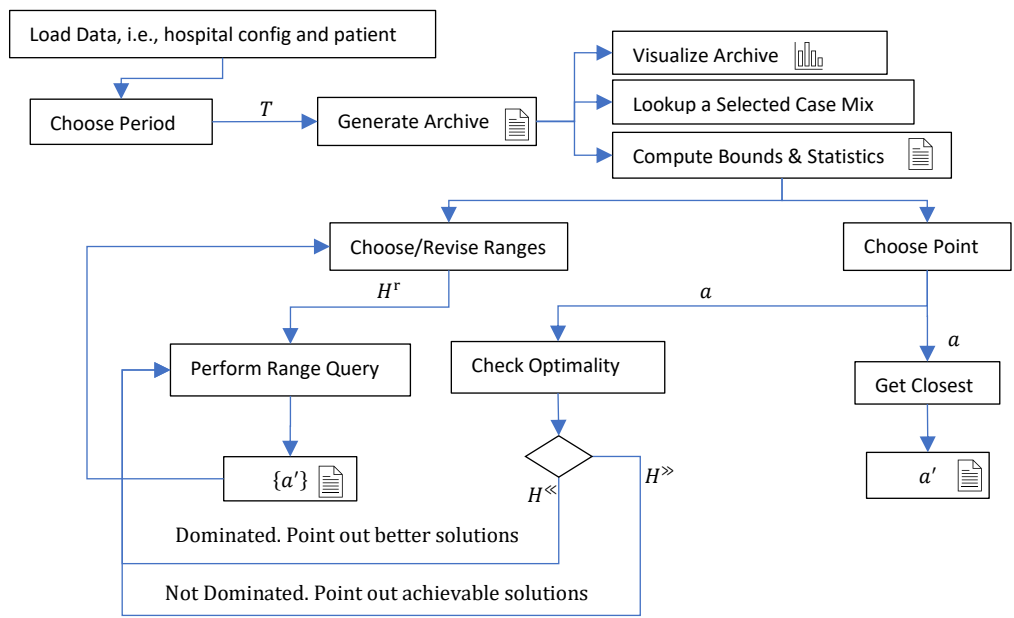

**Figure 6.** Workflow around which a DST can be created.

*Workflow.* The first step after loading hospital and patient data is to choose a time horizon $T$. An appropriate archive of case mix solutions can then be generated using the approaches advocated in earlier sections. If one already exists, it can be loaded. A user should query that archive in two ways. They will specify a case mix solution of interest and investigate the feasibility and optimality of that solution. Otherwise, they will specify a case mix envelope and perform a range query. Revision of these selections should be facilitated by the DST.

*Visuals and Outputs.* Once an archive has been generated, some form of archive visualisation is essential. As the archive can be particularly large, a table including all case mix is undesirable. A hybrid violin chart with box plot (i.e., like Figure 3 and 4), is best suited for summarising an entire archive (i.e., Figure 7a) and describing the exact distribution of values. The violin chart can be shown with raw values, but a normalised view may be more insightful and effective, especially if there is a big discrepancy in the outputs of different groups. To supplement the violin chart, a summary table with lower and upper bounds, and quartiles, should be added. A parallel coordinates chart is an alternative, but in retrospection not well suited as can be seen in Figure 8. Very little information or insight can be seen and there are just too many possibilities. Access to specific case mix should be facilitated. A bar chart or a simple table (i.e., like Figure 7b) is well suited for viewing a particular case mix. In Figure 7b, a slider is present to iterate through the archive members. Range queries and accompanying results can be displayed as shown in Figure 7c. The "View" button permits users to select and view any candidate case mix in more detail using the interface shown in Figure 7b. To check the optimality of a chosen case mix, we suggest an interface like Figure 7d. The "Test" button runs the query and reports the chosen case mix as inferior or else reports it as infeasible. In the former situation, a sub archive of better solutions is extracted and then summarised as a violin chart. In the latter, a sub archive of feasible solutions is instead extracted.

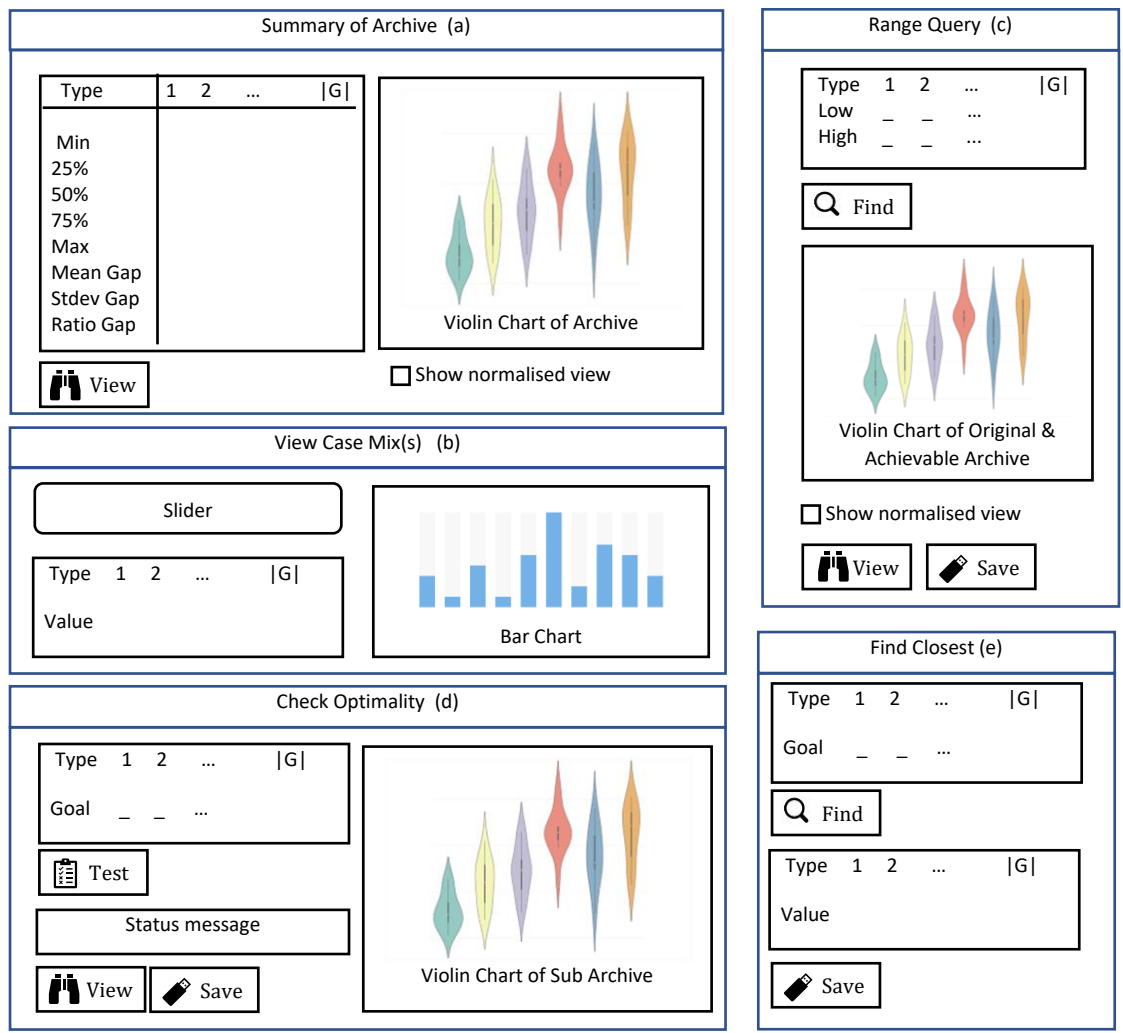

**Figure 7.** Proposed DST

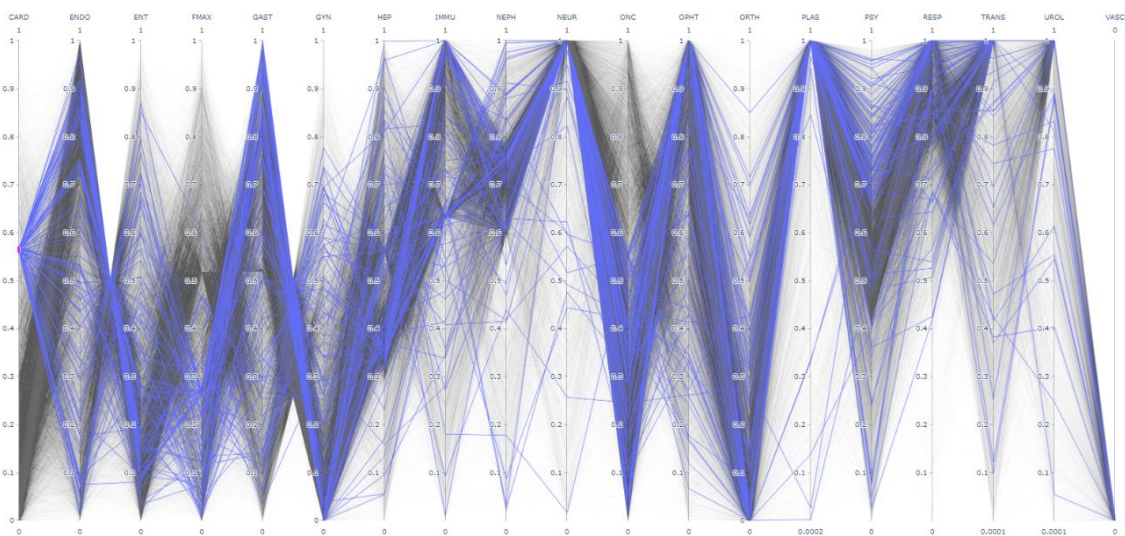

**Figure 8.** Parallel coordinate chart for an archive with 100k points

## 4. Numerical Investigations and Case Study

A numerical investigation is provided in this section to demonstrate the efficacy of the archive generation and querying algorithms. A local tertiary level hospital with 500 plus ward beds, 20 operating rooms, and an ICU with 25 beds is the focus of our case mix sensitivity analysis. The data for 19 patient types is shown in Table 5 and Table 6. A 52-week time horizon was chosen for the analysis. As such, the treatment number limits (i.e., upper bounds) in Table 6 occur.

An archive size of 100,000 solutions was set as a goal of reasonable size to initially evaluate the PRCECM algorithms. In Burdett and Kozan (2016), earlier variants of our algorithms (i.e., RCECM, AECM) were proposed for generating a Pareto frontier in higher dimensions. At that time, the RCECM was able to generate 10,000 non-dominated solutions in about 7.8 hrs. The AECM generated 17,000 solutions and was run for 47hrs before termination. These earlier results provide a reference point to judge the CPU time requirements of our new parallel approaches. In 2016, ILOG Cplex Studio 12.6 was used to solve the different models on a quad core Dell personal computer with 2.6GHz speed and 16GB memory under Windows 7. In contrast, our current numerical testing is run on an 11th Gen i7 with base speed 3.3 GHz (and max speed 4.80GHz), 4 cores (i.e., 8 logical processors), and 16 GB memory under Windows 11. We also used ILOG Cplex Studio 12.10 and concert technology.

**Table 5.** Hospital Infrastructure

| Ward | 1C | 1D | 2A | 2B | 2C | 2D | 2E | 3C | 3D | 3E | 4A | 4BR |
|---|---|---|---|---|---|---|---|---|---|---|---|---|
| #Beds | 24 | 26 | 28 | 26 | 36 | 24 | 29 | 28 | 20 | 14 | 19 | 14 |
| Ward | 4BT | 4C | 4D | 4E | 5A | 5B | 5C | 5D | RENDP | GREV | OT | ICU |
| #Beds | 16 | 28 | 28 | 26 | 28 | 24 | 24 | 24 | 6 | 30 | 19 | 26 |

**Table 6.** Patient types and infrastructure resourcing requirements

| Group | (OT, ICU, WARD) Time (#hrs) | | Ward Used | | Mix | UB |
|---|---|---|---|---|---|---|
| | Inpatient Type (Surg.) | Inpatient Type (Med.) | Inpatient Type (Surg.) | Inpatient Type (Med.) | | |
| CARD | (3.16, 19.85, 171.85) | (0.06, 1.82, 84.45) | 3C | 3D, 3E, 5A | (58.2, 41.2) | 2420.72 |
| ENDO | (2.13, 2.72, 137.85) | (0.51, 0.27, 185.24) | 4D | 4D, 5C | (50.63, 49.37) | 2817.25 |
| ENT | (2.12, 1.02, 44.02) | (0.5, 0.91, 49.43) | 1D | 1D | (54.08, 45.92) | 4884.2 |
| FMAX | (4.52, 6, 131.33) | (0.61, 0.08, 13.55) | 1D | 1D | (70.67, 29.33) | 2346.81 |
| GAST | (2.64, 3.61, 150.71) | (0.144, 0.49, 101.43) | | 4D, 4E, 5C | (54.97, 45.03) | 5301.99 |
| GYN | (2.2, 1.04, 111.36) | (0.59, 0, 52.86) | | 4C, 4E | (67.45, 32.55) | 5109.98 |
| HEPA | (1.475, 4.13, 160.71) | (0.075, 1.84, 119.87) | | 4C, 4E | (45.97, 54.03) | 3402.55 |
| IMMU | (1.93, 4.3, 306.79) | (0.19, 44.68, 149.15) | 2D | 2D, 5B | (5.66, 94.34) | 2652.76 |
| NEPH | (2.19, 0.65, 102.41) | (0.47, 0.143, 50.65) | 4BR | 4BR, RENDP, 5C | (28.3, 71.7) | 4219.99 |
| NEUR | (2.46, 3.67, 243.44) | (0.099, 5.35, 200.68) | 2C | 2C, 5B | (26.95, 73.05) | 2470.08 |
| ONC | (2.86, 2.09, 217.5) | (0.36, 0.89, 172.27) | | 2E | (57.28, 42.72) | 1278.37 |
| OPHT | (1.52, 0.068, 45.35) | (0.046, 0.0, 100.36) | 4D | 5A | (68.33, 31.17) | 7819.4 |
| ORTH | (3.09, 1.93, 218.98) | (0.52, 1.86, 266.12) | | 2A, 2B | (64, 36) | 1999.34 |
| PLAS | (2.43, 1.71, 157.44) | (0.18, 0.1, 137.73) | | 1D | (65.69, 34.31) | 1507.43 |
| PSY | na | (0.08, 0.06, 258.82) | na | GREV | (100) | 1012.6 |
| RESP | (2.86, 3.7, 161.26) | (0.22, 4.76, 136.37) | 2D | 2D, 5A | (5.62, 94.38) | 3297.35 |
| TRANS | (3.33, 445.71, 593.24) | na | 4BT | na | (100) | 235.615 |
| UROL | (1.83, 1.66, 71.63) | (0.38, 0.1, 41.11) | | 4A | (43.73, 56.27) | 3048.02 |
| VASC | (2.98, 4.75, 339.59) | (0.07, 5.9, 122.74) | | 1C | (31.85, 68.15) | 1093.1 |

Both PRCECM algorithms were applied, resulting in the creation of a multitude of archives and violin plot outputs. We chose to evaluate a variety of proximity values of practical relevance. For instance, 0, 50, 100, 200, 500 and 1000, were deemed suitable, where zero implies no proximity testing. Anything larger may be too broad in the context of a hospital case mix assessment. The number of points per thread per stage (i.e., $S$) is expected to influence the PRCECM runtime, so we chose to evaluate six options, leading to the number of stages shown in column 2 of Table 8. Given a time horizon of 52 weeks, the hospital output cannot exceed 56917.55 cases. This is the sum of the upper bounds listed in Table 6. A summary of our numerical testing is shown in Table 7.

**Table 7.** PRCECM parametric results for intended archive size of 100k points

| #PTS_THRD_STAGE | #STAGES | PROX | PRCECM1 | | | PRCECM2 | | |
|---|---|---|---|---|---|---|---|---|
| | | | #GEN. PTS | CPU #HRS | FEAS (%) | #GEN. PTS | CPU #HRS | FEAS (%) |
| **10** | 1250 | 0 | 100000 | 4.91 (3.05) | 1.19 | 100000 | 3.37 (1.95) | 1.26 |
| | | 50 | 99993 | 2.47 (1.93) | 1.23 | 99999 | 3.01 (1.08) | 1.17 |
| | | 100 | 99962 | 5.08 (2.9) | 1.29 | 99987 | 1.89 (2.92) | 1.29 |
| | | 200 | 99749 | 4.80 (1.9) | 1.28 | 99783 | 2.97 (1.87) | 1.29 |
| | | 500 | 90881 | 2.91 (2.89) | 1.32 | 90865 | 2.77 (1.84) | 1.23 |
| | | 1000 | 42855 | 5.21 (2.86) | 1.27 | 43011 | 5.67 (1.13) | 1.28 |
| **50** | 250 | 0 | 100000 | 5.23 | 1.30 | 100000 | 3.11 | 1.21 |
| | | 50 | 100000 | 2.93 | 1.27 | 99999 | 2.53 | 1.30 |
| | | 100 | 99981 | 4.92 | 1.22 | 99974 | 3.08 | 1.18 |
| | | 200 | 99777 | 4.92 | 1.29 | 99710 | 5.49 | 1.25 |
| | | 500 | 90994 | 7.01 | 1.22 | 90864 | 5.64 | 1.21 |
| | | 1000 | 43039 | 5.12 | 1.27 | 43326 | 5.72 | 1.23 |
| **100** | 125 | 0 | 100000 | 4.61 | 1.24 | 100000 | 3.37 | 1.23 |
| | | 50 | 99995 | 4.93 | 1.25 | 99999 | 3.27 | 1.33 |
| | | 100 | 99989 | 4.65 | 1.25 | 99989 | 5.17 | 1.29 |
| | | 200 | 99745 | 4.04 | 1.19 | 99764 | 2.88 | 1.29 |
| | | 500 | 90844 | 4.92 | 1.26 | 91221 | 4.04 | 1.25 |
| | | 1000 | 43055 | 2.84 | 1.29 | 43129 | 3.17 | 1.25 |
| **250** | 50 | 0 | 100000 | 3.33 | 1.31 | 100000 | 1.88 | 1.28 |
| | | 50 | 99997 | 2.43 | 1.26 | 99999 | 4.78 | 1.28 |
| | | 100 | 99980 | 4.02 | 1.22 | 99993 | 4.50 | 1.19 |
| | | 200 | 99751 | 1.59 | 1.24 | 99775 | 3.29 | 1.29 |
| | | 500 | 90980 | 5.14 | 1.24 | 90814 | 2.94 | 1.28 |
| | | 1000 | 42972 | 1.59 | 1.26 | 43156 | 5.03 | 1.27 |
| **500** | 25 | 0 | 100000 | 4.54 | 1.25 | 100000 | 2.04 | 1.24 |
| | | 50 | 99998 | 3.45 | 1.24 | 100000 | 1.71 | 1.24 |
| | | 100 | 99982 | 3.76 | 1.31 | 99985 | 5.00 | 1.21 |
| | | 200 | 99745 | 4.32 | 1.26 | 99764 | 3.14 | 1.29 |
| | | 500 | 91083 | 4.09 | 1.25 | 91183 | 3.51 | 1.30 |
| | | 1000 | 42998 | 5.17 | 1.25 | 43462 | 5.30 | 1.29 |
| **1000** | 12 | 0 | 96000 | 1.60 | 1.35 | 100000 | 1.62 | 1.18 |
| | | 50 | 95996 | 3.52 | 1.18 | 96000 | 3.80 | 1.20 |
| | | 100 | 95981 | 4.41 | 1.28 | 95988 | 2.98 | 1.17 |
| | | 200 | 95782 | 4.37 | 1.21 | 95768 | 4.482 | 1.24 |
| | | 500 | 87516 | 3.77 | 1.21 | 87958 | 3.201 | 1.27 |
| | | 1000 | 41393 | 4.25 | 1.29 | 42681 | 3.62 | 1.32 |
| **12500** | 1 | 0 | 100000 | 4.86 | 1.26 | 100000 | 4.53 | 1.28 |
| | | 50 | 99997 | 4.04 | 1.25 | 100000 | 1.781 | 1.22 |
| | | 100 | 99979 | 4.71 | 1.25 | 99998 | 4.517 | 1.28 |
| | | 200 | 99749 | 1.72 | 1.34 | 99974 | 3.225 | 1.28 |
| | | 500 | 90698 | 1.60 | 1.26 | 98152 | 4.546 | 1.21 |
| | | 1000 | 41867 | 1.78 | 1.22 | 70465 | 1.695 | 1.27 |

Table 7 demonstrates that for the considered 19 objective decision problem, the likelihood of generating a feasible grid point, is quite slim (about 1.3%). As such, we later modified the PRCECM algorithms. The revised and improved version assumes every grid point is infeasible and runs the grid point correction model by default. Given that insight additional testing of a limited nature was performed. In Table 7 the reduced CPU times for selective instances are quoted in brackets. So far, we have seen considerable reductions in CPU time, ranging between 22% and 60% for PRCECM01, and between 33% to 80% for PRCECM02. On only one occasion, was more CPU time required.

***Impact of Proximity Parameter.*** For a specific proximity level, we would expect to see only minor differences in the violin charts produced for different values of the #PTS_THRD_STAGE parameter. Our results show this occurred. Clearly some variation occurs, as different grid points are randomly generated from one run to another, but the main shape is the same.

Overall, the proximity parameter had negligible effect on the violin charts, except when the value was set as 1000. At that value, there are some noticeable differences. An example is shown in Figure 9. The first archive was generated without any proximity restriction and the second with the maximum considered (i.e., 1000). Major differences are explicitly circled. Some of the patient groups have different mean and median, and a wider density, whereas others seem unaffected. During our results analysis we found that the generated points are naturally spaced out, with a Euclidean distance less than 4000 rare. As we only looked at proximity restrictions up to 1000, the PRCECM algorithms rarely need to do any pruning. Hence, that is why most of the violin charts are so similar.

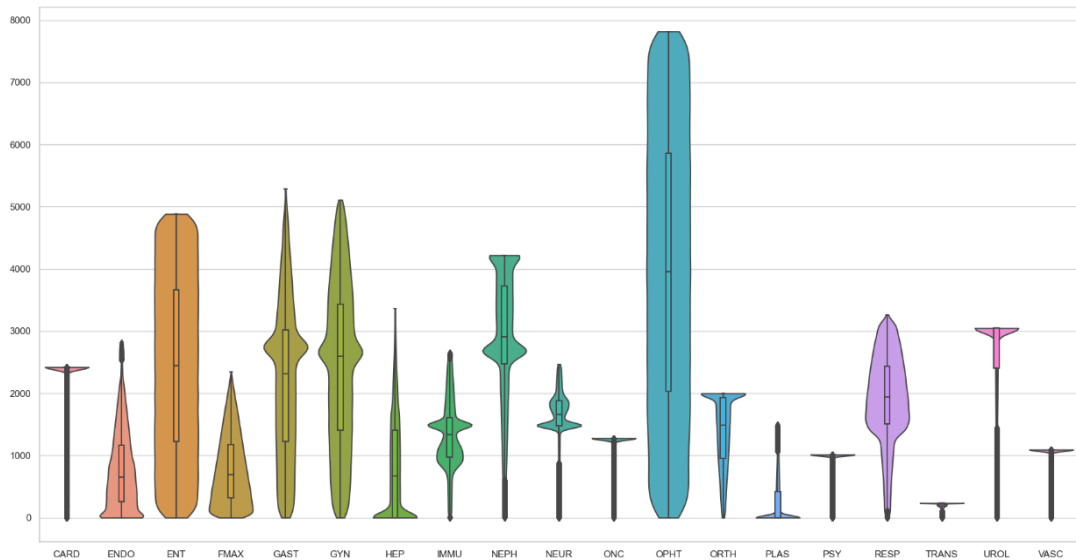

**(a)** No proximity restriction

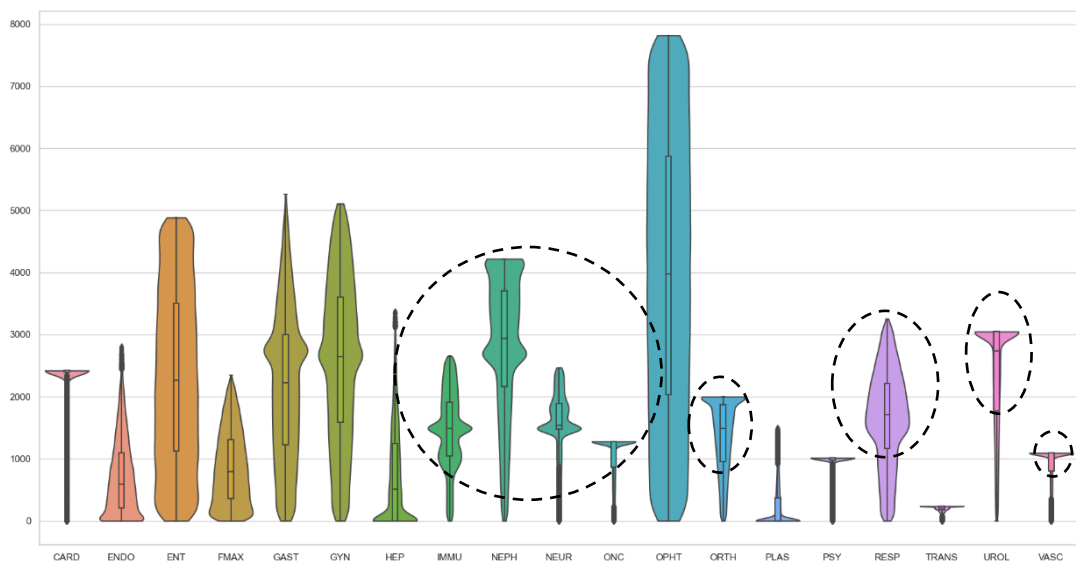

**(b)** Proximity restriction of 1000
**Figure 9.** Violin chart comparison [PRCECM2, #PTS_THRD_STAGE =10]

*CPU Time.* The run time was analysed in more detail and the charts in Figure 10 have been provided. These charts clearly show that PRCECM2 runs faster (i.e., significantly) on most occasions. Figure 10a also shows a slight trend regarding increasing #PTS_THRD_STAGE and reduced CPU time. This was expected. We theorized that if the work done by each thread is larger, then communication with the central archive should be reduced, and less archive updating would be needed. Testing so far has illustrated this to be so. The CPU time, however, is quite variable, depending on the location of the generated grid points within the objective space, and the structure of the KD-Tree upon which case-mix solutions are inserted.

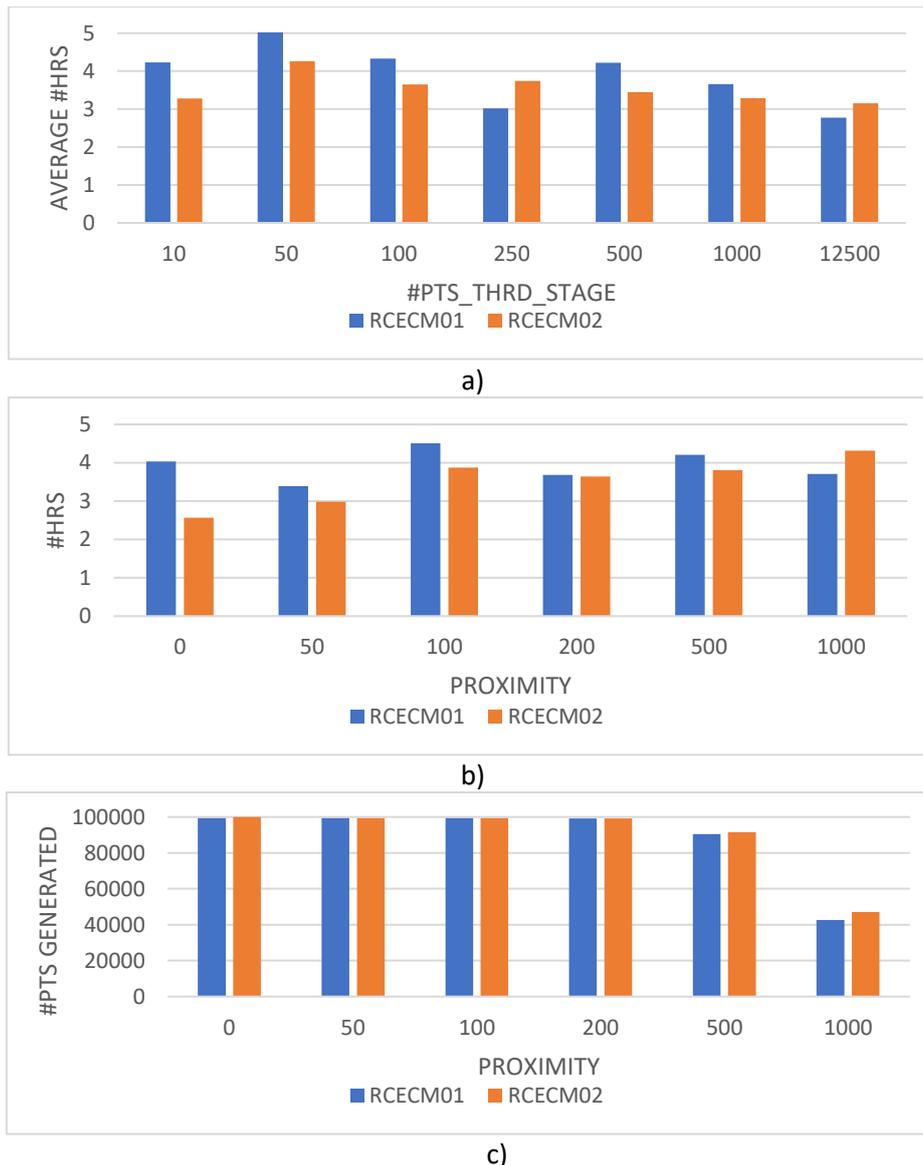

**Figure 10.** Charts related to results in Table 7

There is a definite trend of reduced archive size (Figure 10c) as the proximity parameter is increased. Once the proximity restriction is set to 500 or above, the archive size drops significantly. This is expected because the proximity parameter restricts solutions that are too close to each other. As it is increased more checking is performed and more solutions are likely to be discarded. This inflates runtime. To a lesser extent there appears to be a trend for RCECM02 of increased CPU time (Figure 10b), relative to the proximity parameter.

*Landscape Features*. Looking at the produced violin charts more generally, we can see that the shape of the case mix landscape is quite static. Between different runs, the distribution remains the same. Some patient groups (e.g., CARD, ONC, PSY, TRANS, VASC) have quite narrow distributions. These patient types have dedicated wards and only share operating rooms. As such, trade-offs of a greater nature are less prevalent in the objective space. Some of the other patient groups (e.g., ENT, GAST, GYN, HEPA, OPHT) have a much greater range of outputs. These types have a smaller recovery period, permitting a much higher number of patients to treated over time, if more operating room time is provided. Three of the patient types (e.g., IMMU, NEPH, NEUR) have a more unusual multi-model distribution. Apart from sharing the operating rooms, these types also share wards with other types.

The FMAX, ENDO, and PLAS types have a distinctive triangular looking distribution with highest representation at lower levels of output.

Normalizing the data before plotting violin charts is also worthwhile and insightful. The violin chart in Figure 9a for instance is equivalent to Figure 12. The y-axis is the normalized treatment numbers.

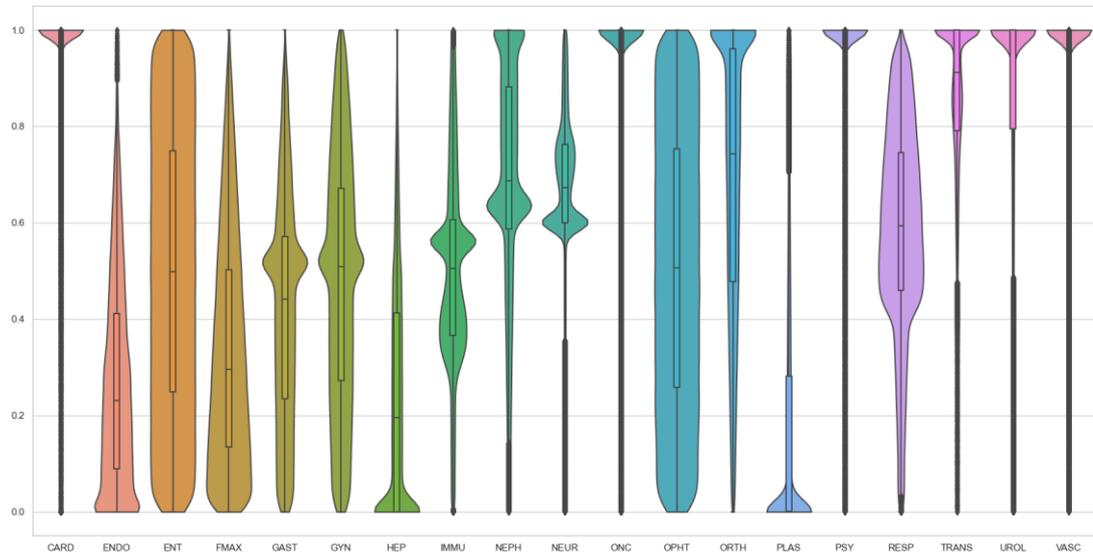

**Figure 12.** Normalized Violin chart

A violin plot of the total patients treatable is shown in Figure 13. Many case mix solutions exist with high output, but the range is wide, and low outputs also selectable.

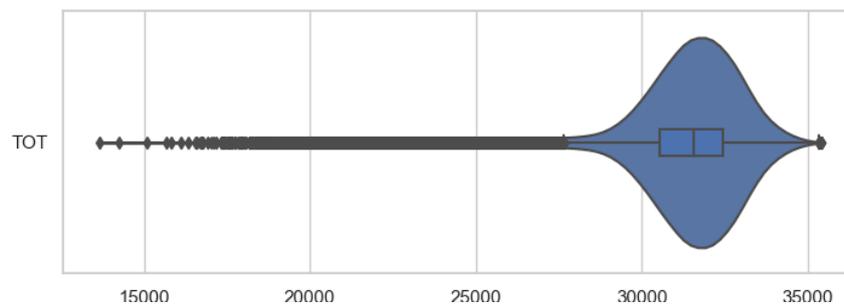

**Figure 13.** Total patients treatable

*Archive Uniformity.* Each of the generated archives has in each dimension a specific spread of values. Under scrutiny, the spread appears quite uniform in each archive as indicated by small mean gap size and standard deviation. In Figure 14 values associated with two selected archives are shown. These charts demonstrate that the spread is affected by the proximity setting used. At larger proximity settings the gap is larger. The coefficient of variation (i.e., the ratio), is above one, a sign that the variations in the gap are not always minor. This is also substantiated up by the maximum gap observed.

*Generating Larger Archives.* Our numerical testing of the PRCECM algorithms initially involved the generation of archives of 100,000 points. The archive generation time that we witnessed of about 3-5 hrs is reasonable for health care managers and planners given access to standard computing resources such as laptops and other personal computers. Our numerical testing captured principal characteristics, but there was evidence to suggest archive uniformity could be improved. This, however, requires more points to be generated. To further improve our understanding of the real frontier, much larger archives were generated using a High-Performance Computing environment with 146 compute nodes, and 7640 CPU cores (i.e., Intel Xeon Gold 6140, AMD EPYC 7702). The results are summarised in Table 8.

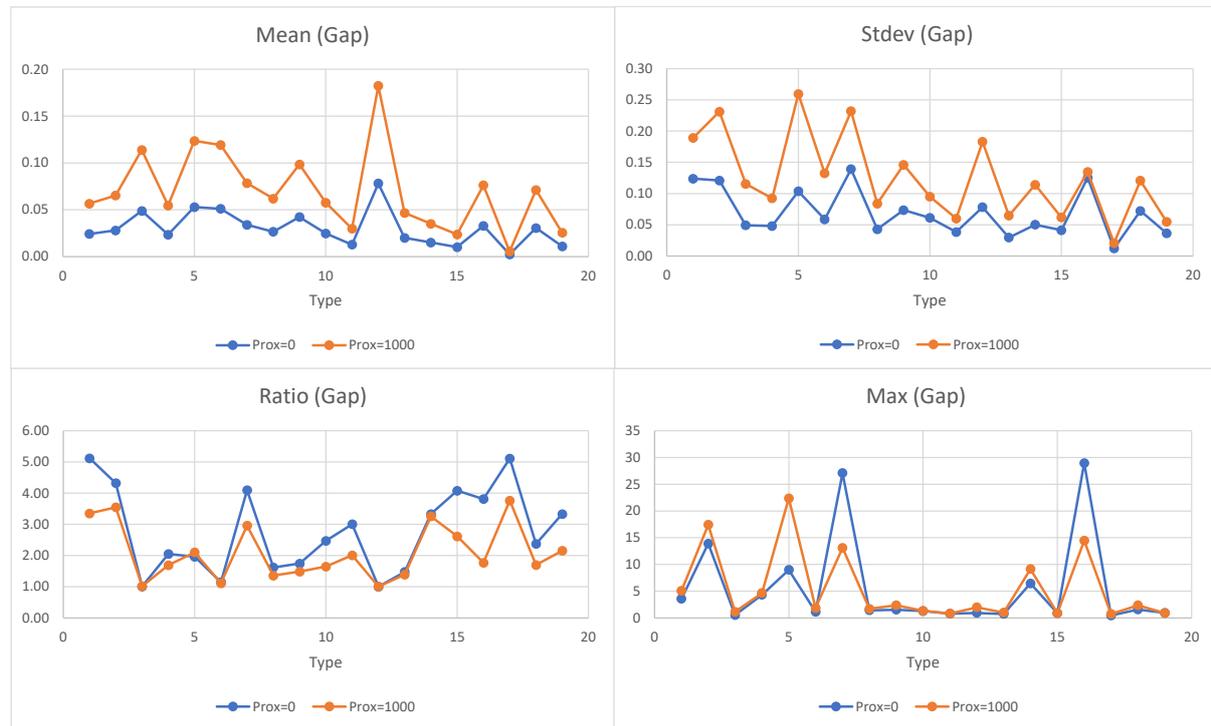

**Figure 14.** Archive gap information (#PTS_THRD_STAGE=10|PRCECM01)

Interestingly, the larger archives produced have similar violin charts, and there are no noticeable differences to those shown earlier in this article. However, the uniformity and detail is better. This observation is substantiated by viewing different two- and three-dimensional scatter plots (i.e., projections) of the archive's solutions. A selection of those is shown in Appendix E and F. In the right-hand corner are noticeable blank areas where trade-offs restrict certain case mix solutions from occurring. Many of these charts show interesting bands, where the density of points is greater. Some regions evidently have a greater number of case mix solutions, but these can only be seen in other dimensions.

**Archive Querying**. Relative to our case study, several range and goal queries are demonstrated in this section. The range query in Table 9 is a specific case of a general [goal, upper bound] range query. Queries of this nature identify solutions exceeding a specified minimum level of output (a.k.a., achievement) for each patient type. On this occasion, the goal is 25% of the upper bound. Of the 100,000 solutions in the queried archive (i.e., [alg=1, #PTS_THRD_STAGE=10, PROX=0]), 89 satisfied the specified conditions. The details of those are summarised in Figure 16b using violin charts. For brevity, only two violin charts are shown. More detailed comparisons can be made, by looking at individual patient types (e.g., comparing original, requested and achievable), but that would necessitate the display of 19 separate violin charts.

Range queries of the form [0, goal] are unproductive, in the sense that case mix with low outputs are of little interest. Furthermore, the existence of a Pareto optimal case mix with $n_g \leq goal_g \ \forall g \in G$ are unlikely, except when the goal is a high percentage of the upper bound. For example, with the goal set at 90%, 1186 case mix are identifiable. At 80%, only 14 exist.

In Table 10, some goals were arbitrarily defined, and an optimality check was then performed. This goal was found to be achievable. In the consulted archive, we found 699 superior case mix solutions. These are summarised in Figure 17, where the goal is represented by the lower line. In Table 11, goals later found to be unachievable were also queried. No case mix solutions were found in the dominance region of the specified goal, so we proceeded to look for the closest case mix in the archive. This is also shown in the lower line of Table 11.

**Table 8.** Results of larger numerical tests

| #PTS REQ | #PTS_THRD_STAGE | PROX | PRCECM1 | | PRCECM2 | |
|---|---|---|---|---|---|---|
| | | | #GEN. PTS | CPU #HRS | #GEN. PTS | CPU #HRS |
| **200,000** | 1000 | 0 | 200,000 | 1.74 | 200,000 | 1.58 |
| | | 50 | 199,996 | 1.55 | 199,992 | 2.96 |
| | | 100 | 199,926 | 1.56 | 199,918 | 1.55 |
| | | 200 | 199,067 | 1.66 | 199,059 | 1.98 |
| | | 500 | 172,361 | 1.66 | 172,756 | 1.00 |
| | | 1000 | 70,722 | 1.55 | 71,508 | 1.13 |
| | 5000 | 0 | 200,000 | 1.58 | 200,000 | 1.53 |
| | | 50 | 199,994 | 1.58 | 199,992 | 1.02 |
| | | 100 | 199,930 | 1.79 | 199,934 | 1.41 |
| | | 200 | 199,110 | 1.64 | 199,175 | 1.03 |
| | | 500 | 172,386 | 1.56 | 175,217 | 1.11 |
| | | 1000 | 70,330 | 1.63 | 77,820 | 1.60 |
| **300,000** | 1000 | 0 | 296,001 | 3.00 | 296,000 | 1.98 |
| | | 50 | 295,985 | 2.87 | 295,988 | 2.95 |
| | | 100 | 295,846 | 2.95 | 295,856 | 2.59 |
| | | 200 | 294,065 | 2.71 | 294,137 | 2.35 |
| | | 500 | 245,585 | 2.92 | 246,370 | 2.75 |
| | | 1000 | 93,704 | 3.03 | 94,699 | 1.81 |
| | 5000 | 0 | 280,000 | 2.49 | 280,000 | 3.06 |
| | | 50 | 279,989 | 2.57 | 279,985 | 2.80 |
| | | 100 | 279,873 | 2.12 | 279,886 | 2.80 |
| | | 200 | 278,256 | 2.60 | 278,428 | 1.87 |
| | | 500 | 233,563 | 2.59 | 236,826 | 1.10 |
| | | 1000 | 89,587 | 2.57 | 96,427 | 2.72 |
| **400,000** | 1000 | 0 | 400,000 | 5.50 | 400,000 | 5.21 |
| | | 50 | 399,974 | 4.84 | 399,983 | 5.22 |
| | | 100 | 399,701 | 5.83 | 399,772 | 6.25 |
| | | 200 | 396,667 | 1.97 | 396,699 | 5.09 |
| | | 500 | 321,194 | 2.07 | 322,019 | 2.52 |
| | | 1000 | 116,541 | 2.44 | 117,401 | 2.55 |
| | 5000 | 0 | 400,000 | 4.36 | 400,000 | 6.11 |
| | | 50 | 399,982 | 4.6 | 399,979 | 4.89 |
| | | 100 | 399,698 | 4.92 | 399,783 | 4.75 |
| | | 200 | 396,566 | 5.3 | 396,982 | 4.69 |
| | | 500 | 320,890 | 5.71 | 324,584 | 5.29 |
| | | 1000 | 115,208 | 4.9 | 122,971 | 5.12 |
| **500,000** | 1000 | 0 | 496,000 | 5.23 | 496,000 | 8.51 |
| | | 50 | 495,968 | 5.88 | 495,970 | 7.90 |
| | | 100 | 495,599 | 3.42 | 495,593 | 8.65 |
| | | 200 | 491,084 | 8.96 | 491,041 | 8.10 |
| | | 500 | 388,280 | 9.30 | 388,852 | 9.20 |
| | | 1000 | 135,685 | 8.74 | 136,715 | 9.22 |
| | 5000 | 0 | 480,000 | 7.84 | 480,000 | 2.04 |
| | | 50 | 479,968 | 7.75 | 479,975 | 7.41 |
| | | 100 | 479,626 | 7.69 | 479,658 | 7.94 |
| | | 200 | 475,326 | 7.92 | 475,439 | 7.79 |
| | | 500 | 377,007 | 3.49 | 380,468 | 8.10 |
| | | 1000 | 132,188 | 5.16 | 139,029 | 2.80 |

**Table 9**. A demonstrative range query

| Group | CARD | ENDO | ENT | FMAX | GAST | GYN | HEPA | IMMU | NEPH | NEUR |
|---|---|---|---|---|---|---|---|---|---|---|
| Low | 806 | 248 | 1220 | 313 | 1248 | 1402 | 850 | 971 | 2494 | 1482 |
| High | 2420.7 | 2817.25 | 4884.2 | 2346.81 | 5301.99 | 5109.98 | 3402.55 | 2652.76 | 4219.99 | 2470.08 |

| Group | ONC | OPHT | ORTH | PLAS | PSY | RESP | TRANS | UROL | VASC |
|---|---|---|---|---|---|---|---|---|---|
| Low | 320 | 2032 | 960 | 377 | 253 | 1504 | 186 | 2432 | 274 |
| High | 1278.37 | 7819.4 | 1999.34 | 1507.43 | 1012.6 | 3297.35 | 235.62 | 3048.02 | 1093.1 |

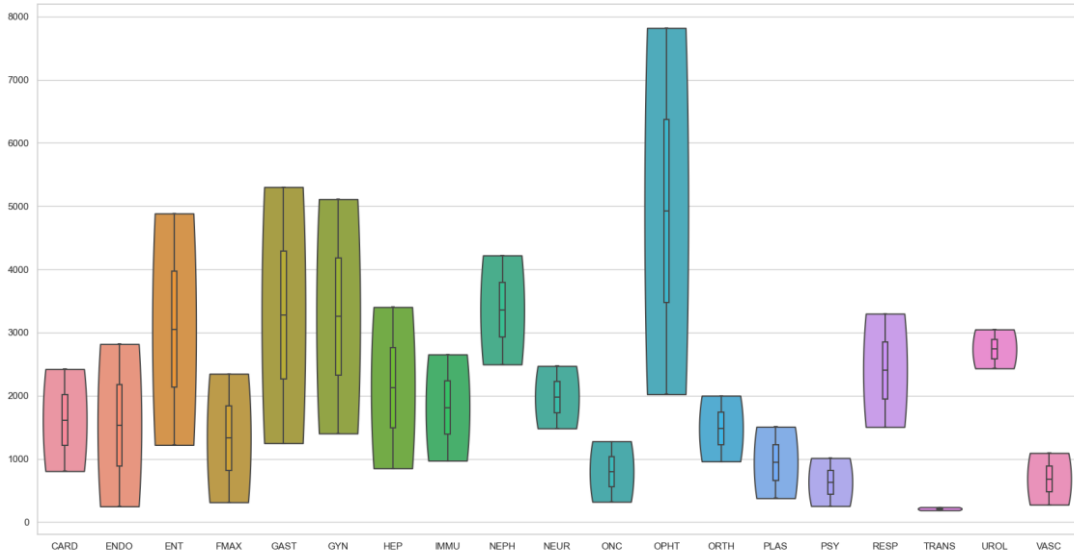

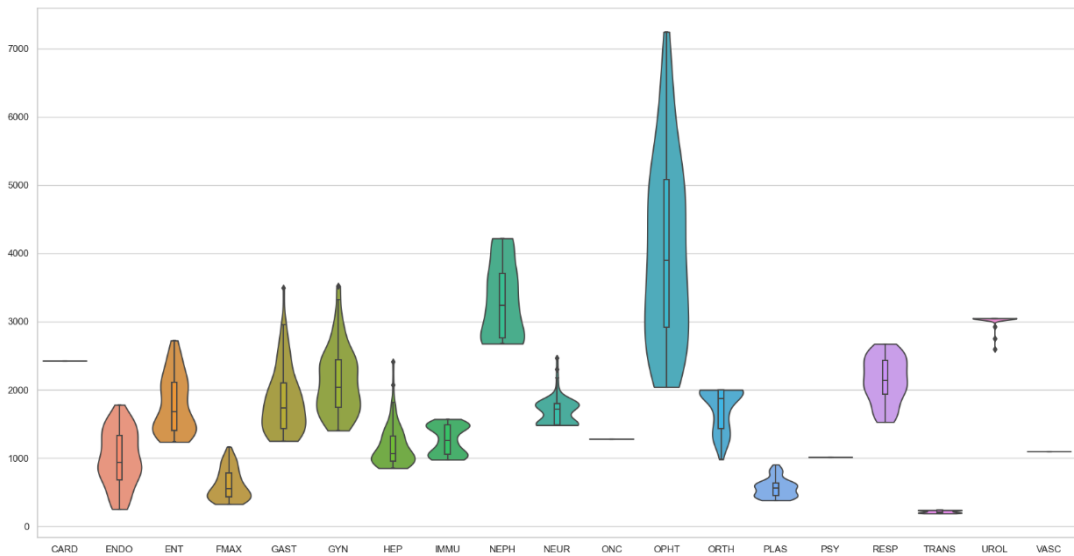

**Figure 16**. Summary of range query. a) Requested b) Achieved

**Table 10.** A demonstrative set of goals

| Group | CARD | ENDO | ENT | FMAX | GAST | GYN | HEPA | IMMU | NEPH | NEUR |
|---|---|---|---|---|---|---|---|---|---|---|
| Goal | 1157 | 320 | 1222 | 344 | 1167 | 1297 | 600 | 667 | 1450 | 800 |

| Group | ONC | OPHT | ORTH | PLAS | PSY | RESP | TRANS | UROL | VASC |
|---|---|---|---|---|---|---|---|---|---|
| Goal | 581 | 1974 | 689 | 122 | 478 | 957 | 102 | 1308 | 505 |

**Table 11.** A demonstrative set of unachievable goals and the closest optimal case mix

| Group | CARD | ENDO | ENT | FMAX | GAST | GYN | HEPA | IMMU | NEPH | NEUR |
|---|---|---|---|---|---|---|---|---|---|---|
| Goal | 2315 | 641 | 2444 | 689 | 2335 | 2594 | 1200 | 1334 | 2900 | 1600 |
| Closest | 2420.72 | 836.182 | 2088.73 | 884.143 | 2198.94 | 2045.6 | 1197.52 | 1490.06 | 3177.4 | 1482.05 |
| Change | 105.68 | 195.182 | -355.27 | 195.143 | -136.06 | -548.4 | -2.48 | 156.06 | 277.4 | -117.95 |

| Group | ONC | OPHT | ORTH | PLAS | PSY | RESP | TRANS | UROL | VASC |
|---|---|---|---|---|---|---|---|---|---|
| Goal | 1162 | 3948 | 1379 | 244 | 956 | 1915 | 204 | 2617 | 1011 |
| Closest | 1278.37 | 3874.62 | 1481.54 | 294.86 | 1012.6 | 2229.76 | 189.78 | 3048.02 | 1093.1 |
| Change | 116.37 | -73.38 | 102.54 | 50.862 | 56.6 | 314.76 | -14.22 | 431.02 | 82.1 |

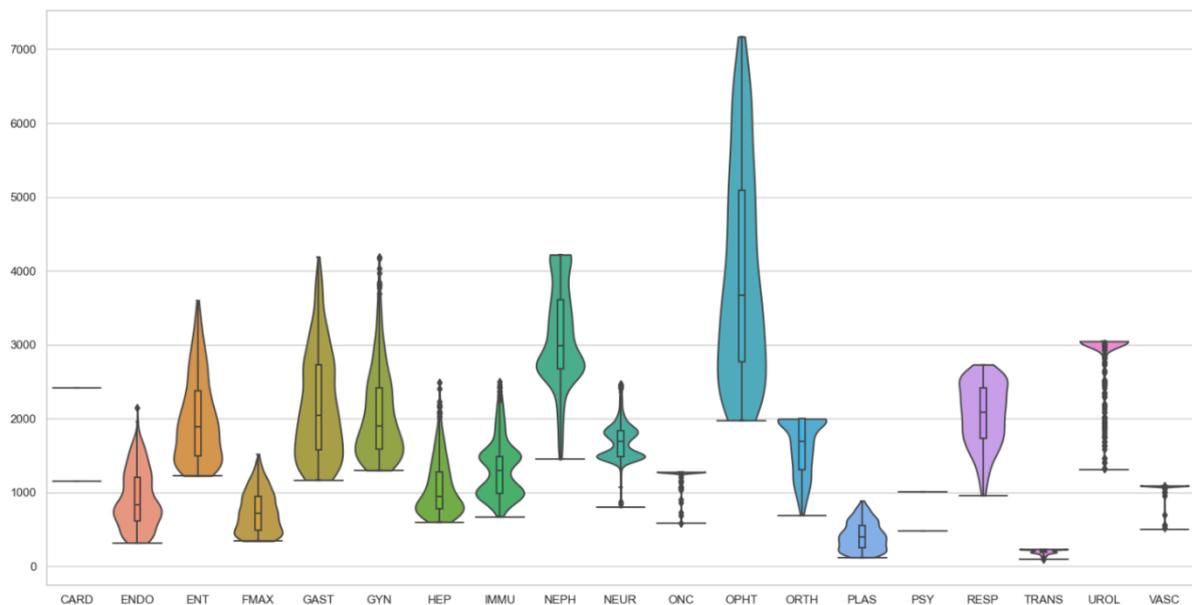

**Figure 17.** Solutions better than the designated goal case mix

## 5. Conclusions

It is worthwhile for healthcare managers and planners to better understand the case mix landscape of their hospital, so that they can better manage their hospital activities, resources, revenues, and costs. Identifying the case mix landscape is analogous to generating an archive of Pareto optimal case mix. As this task is computationally difficult, we have proposed and tested a new multicriteria optimization approach. Our numerical testing demonstrates that a sufficiently large archive can be generated in a brief period using this article's proposed parallel random corrective epsilon constraint method. As we have integrated a KD-Tree data structure to efficiently store an archive, this permits us to perform additional comparison checking. Therefore, we can eliminate case mix solutions that are not too dissimilar to others and generate a more homogenous frontier.

The premise of the parallel random corrective epsilon constraint method is to iteratively evaluate many randomly chosen grid points in parallel, coalescing those at interims into a central archive. If the generated grid point is infeasible, a corrective model is applied to obtain a feasible point from which a Pareto optimal solution can be extracted. Our numerical testing indicates that most points are infeasible, so the corrective model should be applied upfront. The effect of this modification is reduced algorithm CPU time, as unnecessary model solves are eliminated.

Once generated, a case mix archive can be queried to provide answers to a variety of weight capacity and output related questions. This article has demonstrated through the development of a decision support tool how different forms of query can be posed and answered.

KD-Trees are used in several ways in this article. In future research, it is worth considering whether a KD-Tree can be used to flush out regions that are worthy of further investigation. Currently grid points are selected randomly without any understanding of the objective space.

**Acknowledgements:** This research was funded by the Australian Research Council (ARC) Linkage Grant LP 180100542 and supported by the Princess Alexandra Hospital and the Queensland Children's Hospital in Brisbane, Australia.

**Appendix A: List Based Archive Implementation (Selected Procedures)**

Archive :: RangeQuery($\mathcal{H}^r$):
    1. $\mathcal{C} \leftarrow \text{Filter}(\mathcal{H}^r)$; // Identify achievable solutions
    2. return $\mathcal{C}$

Archive :: RangeQueryExt($\mathcal{H}^r$):
    1. $\mathcal{C} \leftarrow \text{RangeQuery}(\mathcal{H}^r)$; // Identify achievable solutions
    2. if($|\mathcal{C}| > 0$) { $\mathcal{H}^a \leftarrow \text{ComputeBounds}(\mathcal{C})$; $a^* \leftarrow \text{GetNearestNeighbour}(a^I, \mathcal{C})$; } // Compute hypercube
    3. else { $\mathcal{C} \leftarrow \emptyset$; $a^* \leftarrow NULL$; $\mathcal{H}^a \leftarrow NULL$; }
    4. return $(\mathcal{C}, a^*, \mathcal{H}^a)$;

Archive :: ComputeBounds( ):
    1. return $\mathcal{H} \leftarrow \{(lb_k, ub_k)\}_{k \in \{1,...,K\}}$ where $lb_k \leftarrow \min_{a \in \mathcal{A}}(a_k)$ and $ub_k \leftarrow \max_{a \in \mathcal{A}}(a_k)$;

Archive :: IsIn $(a^r) \mapsto (bool)$
    1. $\forall a \in \mathcal{A}$: if(same_point($a^r, a$)) return true;
    2. return false;

Archive :: Filter($\mathcal{H}$):
    1. $\mathcal{C} \leftarrow \emptyset$;
    2. $\forall a \in \mathcal{A}$: if(InRange($a, \mathcal{H}$)) $\mathcal{C} \leftarrow \mathcal{C} \cup \{a\}$; // Enlarge set $\mathcal{C}$
    3. return $\mathcal{C}$;

Archive :: InRange($a, \mathcal{H}$):

1. $i \leftarrow 0$;
2. do { if($a_k < lb_k \lor a_k > ub_k$) return *false*; else $k \leftarrow k + 1$; } while ($k < K$);
3. return *true*;

Archive :: GetClosest($a^r$): return $\arg\min_{a \in \mathcal{A}} \|a - a^r\|$;

Archive :: FindMin($k$): return $\arg\min_{a \in \mathcal{A}}(a_k)$;

Archive :: FindMax($k$): return $\arg\max_{a \in \mathcal{A}}(a_k)$;

Archive :: GetNeighbours($a, radius, \mathcal{C}$):
1. $\forall a' \in \mathcal{A}$: if($\|a - a'\| \leq radius$) $\mathcal{C} \leftarrow \mathcal{C} \cup \{a'\}$;

Archive :: IsDominated($a$):
1. $\forall a' \in \mathcal{A}$: if$\big(dominates(a', a)\big)$ return true;
2. return false;

Archive :: dominates($a, a'$):
1. $better = 0$;
2. $\forall k \in K$:
   if($a_k > a'_k$) $better \leftarrow better + 1$;
   if($a_k < a'_k$) return false; // Worse, so can't be non-dominated
3. return ($better > 0$);

Archive :: FindNonDominated($\mathcal{PF}$): Partition$\big(\mathcal{A}^{\text{non}}, \mathcal{A}^{\text{dom}}\big)$; $\mathcal{PF} = \mathcal{A}^{\text{non}}$;

Archive :: AnalyseUniformity($\mathcal{A}$):
1. $\forall k \in K$: AnalyseGaps($k, \mathcal{A}, \mu_k, \sigma_k$); // Implemented as a parallel for loop

Archive :: AnalyseGaps($k, \mathcal{A}, \mu_k, \sigma_k$):
1. sort($\mathcal{A}, k$); // Sort points in ascending order by kth value
2. $\mu_k \leftarrow 0$; $\sigma_k \leftarrow 0$;
3. $\forall n \in \{1, \ldots, |\mathcal{A}| - 1\}$:
   $gap_k[n] \leftarrow a_k[n+1] - a_k[n]$;
   $\mu_k \leftarrow \mu_k + (1/n)(gap_n - \mu_k)$; // Recursive mean calculation
   $\sigma_k \leftarrow [(n-1)/n] \times (\sigma_k + (1/n)(gap_n - \mu_k)^2)$; // Recursive variance calculation

Archive :: AnalyseSpread($\mathcal{A}$):
1. $\forall k \in K$: // Implemented as a parallel for loop
   $\mathcal{A}_k = (a_k[n])_{n \in \{1 \ldots |\mathcal{A}|\}}$; // An array of the dimension $k$ values
   $\mu_k \leftarrow$ ComputeMean($\mathcal{A}_k$);
   $(q_k^1, q_k^2, q_k^3) \leftarrow$ ComputeQuartiles($\mathcal{A}_k$); // 1st, 2nd and 3rd quartiles
2. return $(\mu, q^1, q^2, q^3)$;

**Appendix B: KDT Based Archive Implementation (Selected Procedures)**

Archive :: RangeQuery($\mathcal{H}^r$):
1. $\mathcal{T}.\text{range\_query}(\mathcal{H}^r, \mathcal{C})$;
2. return $\mathcal{C}$

Archive :: ComputeBounds( ):
1. return $\mathcal{H} \leftarrow \big\{\big(\mathcal{T}.\text{find\_min}(k), \mathcal{T}.\text{find\_max}(k)\big)\big\}_{k \in \{1, \ldots, K\}}$ // Parallel for loop

Archive :: Filter($\mathcal{H}$):
1. $\mathcal{C} \leftarrow \emptyset$;
2. $\mathcal{T}.\text{range\_query}(\mathcal{H}, \mathcal{C})$;
3. return $\mathcal{C}$;

Archive :: IsIn ($a^r$): return $\mathcal{T}.\text{is\_in}(a^r)$;
Archive :: GetClosest: ($a^r$): return $\mathcal{T}.\text{get\_nearest\_neighbour}(a^r)$;
Archive :: FindMin($k$): return $\mathcal{T}.\text{find\_min}(k)$;
Archive :: FindMax($k$): return $\mathcal{T}.\text{find\_max}(k)$;
Archive :: GetNeighbours($a, radius, \mathcal{C}$): $\mathcal{T}.\text{get\_neighbours}(a, radius, \mathcal{C})$;
Archive :: IsDominated($a$): return $\mathcal{T}.\text{is\_dominated}(a)$;
Archive :: FindNonDominated($\mathcal{PF}$): $\mathcal{T}.\text{find\_non\_dominated}(\mathcal{A}, \mathcal{PF})$;

**Appendix C: Selected KDT Implementation**

KDT::is_dominated($a$):
1. $\forall k \in K: \mathcal{H}_k \leftarrow (a_k, ub_k)$; // Note: $ub_k = \mathcal{T}.\text{find\_max}(k)$
2. return is_dominated($a, root, \mathcal{H}, 0$);

KDT::is_dominated($a, node, \mathcal{H}, depth$):
1. if(in_range($a, \mathcal{H}$) and ¬same_point($a, node.pt$)): return true;
2. $cd = depth \% K$; // Modulo operation
3. if($node.left \neq NULL$ and $node.pt_k > \mathcal{H}.lb_k$)
   if$\bigl(is\_dominated(a, node.left, \mathcal{H}, depth + 1)\bigr)$ return true;
4. if($node.right \neq NULL$ and $node.pt_k \leq \mathcal{H}.ub_k$)
   if$\bigl(is\_dominated(a, node.right, \mathcal{H}, depth + 1)\bigr)$ return true;
5. return false;

KDT::find_non_dominated($\mathcal{A}, \mathcal{PF}$):
1. $\forall a \in \mathcal{A}: flag_a \leftarrow is\_dominated(a)$; // Parallel for loop
2. $\forall a \in \mathcal{A}: if\bigl(flag_g\bigr) \mathcal{PF} \leftarrow \mathcal{PF} \cup \{a\}$; // Standard for loop

**Appendix D.** Example with 30 points
Archive: [25,5,87], [28,74,50], [65,47,14], [76,45,25], [79,90,8], [10,66,17], [100,89,82], [96,15,33], [49,64,52], [84,30,47], [97,33,9], [68,26,96], [68,93,76], [12,95,13], [98,35,42], [98,33,1], [61,31,25], [26,66,50], [58,6,75], [50,61,31], [9,11,33], [19,54,47], [11,62,2], [44,89,49], [27,5,41], [38,81,29], [80,79,78], [51,28,31], [46,88,4], [42,62,36].
Requested: [45,20,56], [100,95,96]
Achievable: [100,89,82],[68,26,96],[68,93,76],[80,79,78]

**Appendix E. Trade-offs Between Specialties (2D Projections)**

In this Appendix, various 2-dimensional projections of the case mix landscape are shown. Each point represents a different case mix solution, and each axis describes the number of patients treated of a specific specialty. Regions of higher density, visible as darker areas, highlight the presence of planes in other dimensions. Some of these planes are visible in the 3-dimensional plots shown in Appendix F.

Gynaecology versus Hepatology. Both specialties share ward 4C and 4E, and as such, the trade-off region (i.e., top right side) is quite large. The dense banded region parallel to the diagonal is distinctive.

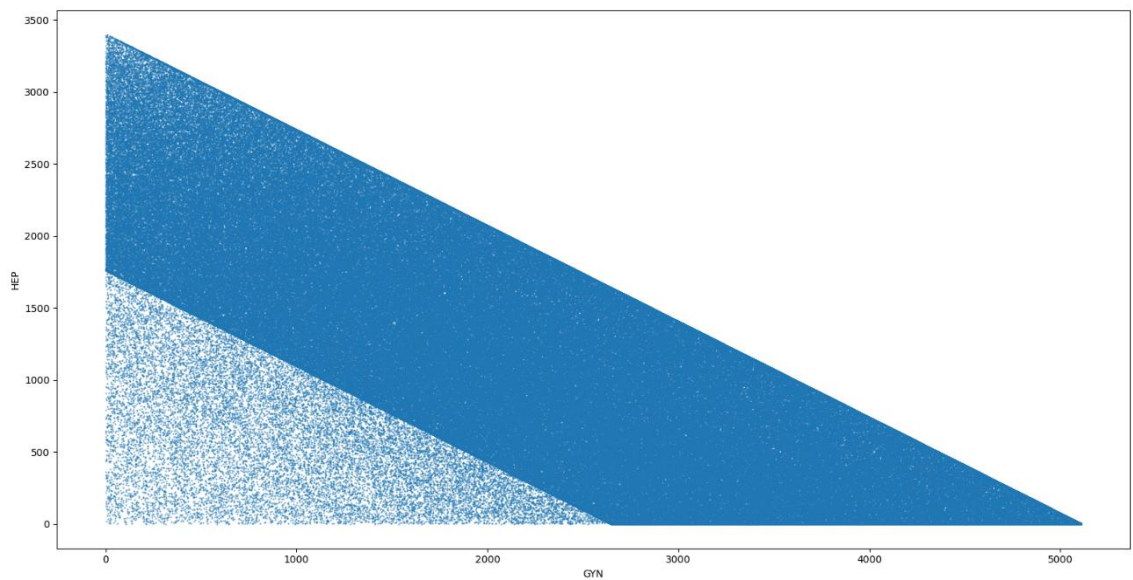

Immunology versus Neurology. These specialties only share ward 5B. As such the trade-off region is quite small. The trade-off region does not originate at the anchor points, e.g., the maximum value of Neurology can occur for varying levels of Immunology and vice versa. The landscape is quite interesting and complex from this projection. There are several unique sub regions, and the density of solutions in those regions is different.

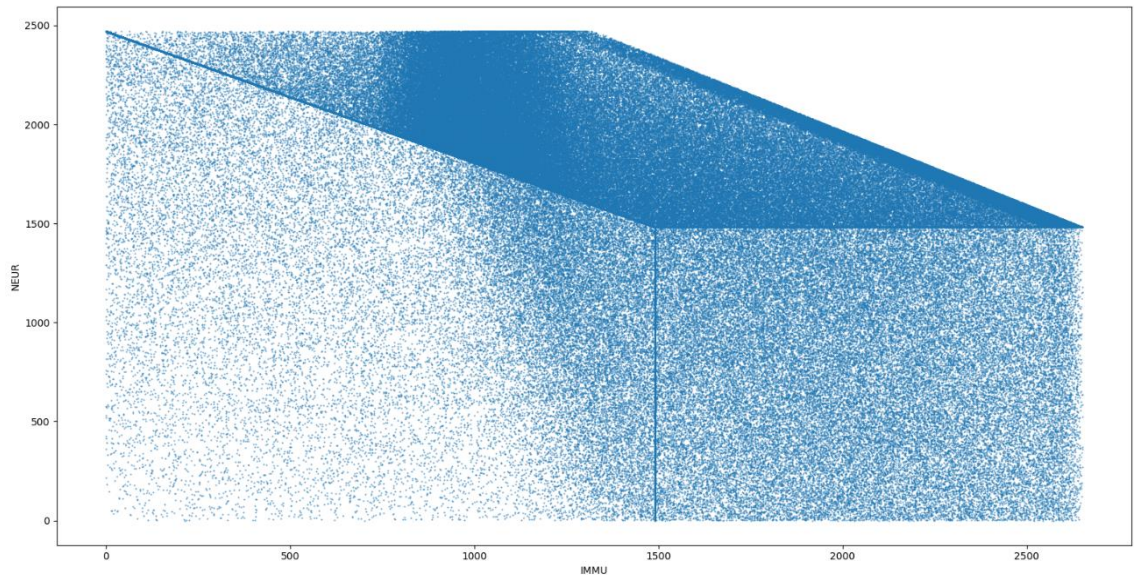

Immunology versus Respiratory. These specialties only share ward 2D. Similar comments can be made about the trade-off region. The banded structure is quite complex and distinctive.

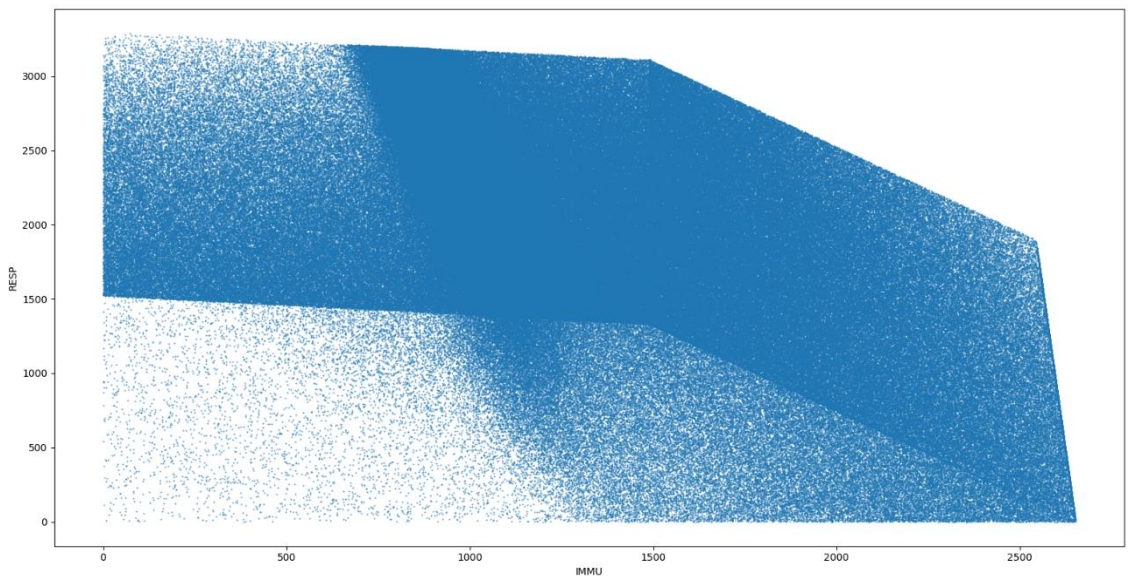

Gynaecology versus Gastroenterology. These specialties share ward 4E. The landscape is quite uniformly distributed, but there are some visible and distinct bands indicating the presence of planes.

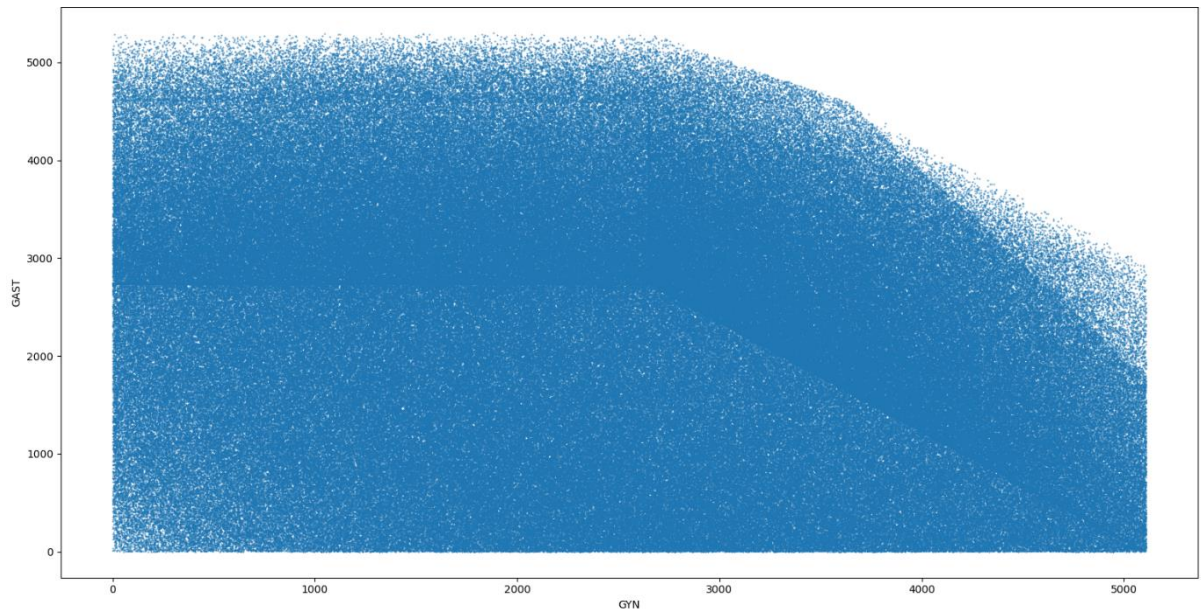

Endocrine versus Nephrology. These specialties share ward 5C. The upper left corner is quite dense while the right-hand side is not. Several horizontal and vertical planes are evident.

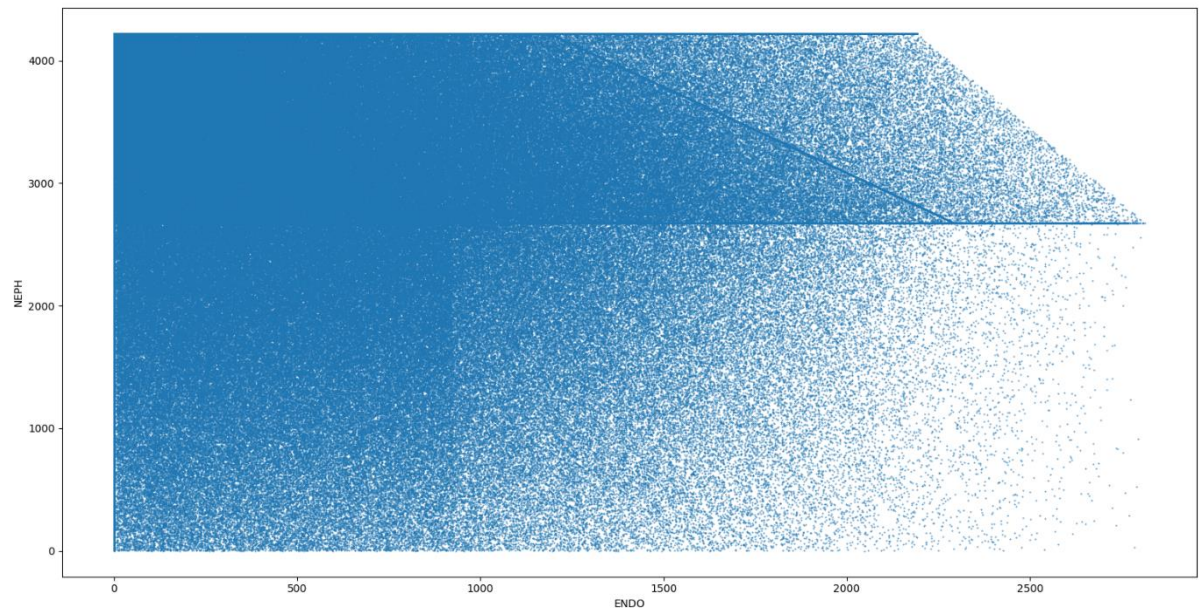

Endocrine versus Gastroenterology. These specialties share ward 4D and 5C and as such, the trade-off region is larger. There is a very heavily banded region midway, and a lighter one above that.

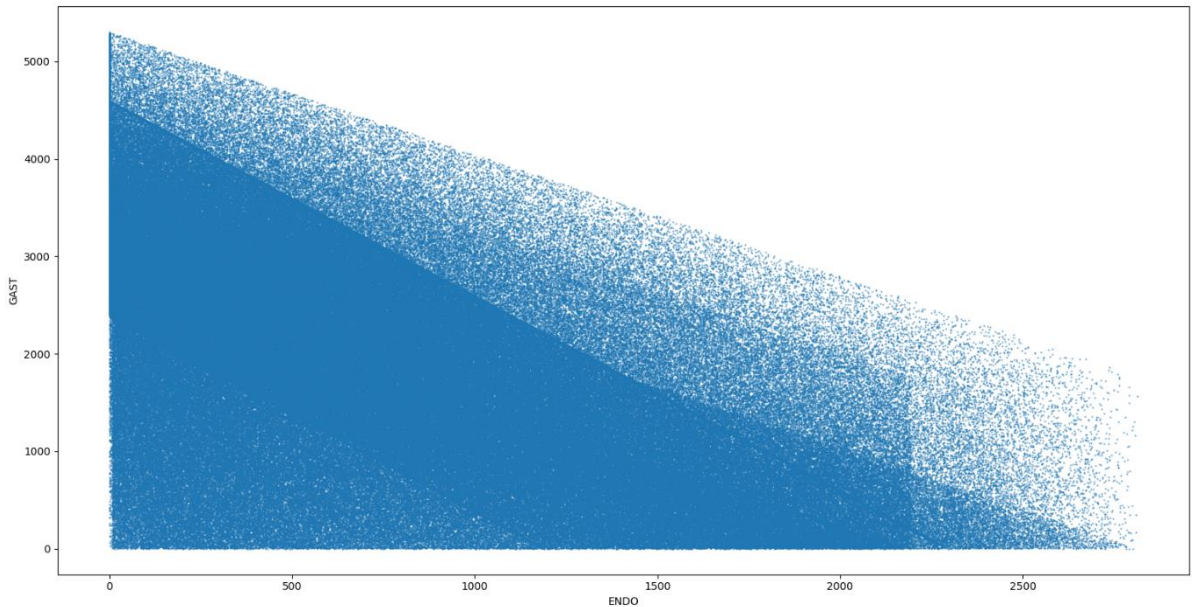

Endocrine versus Ophthalmology. These specialties share ward 4D. The trade-off region is very large and is piecewise linear. There are two distinctive diagonal lines and a less dense region at the bottom right.

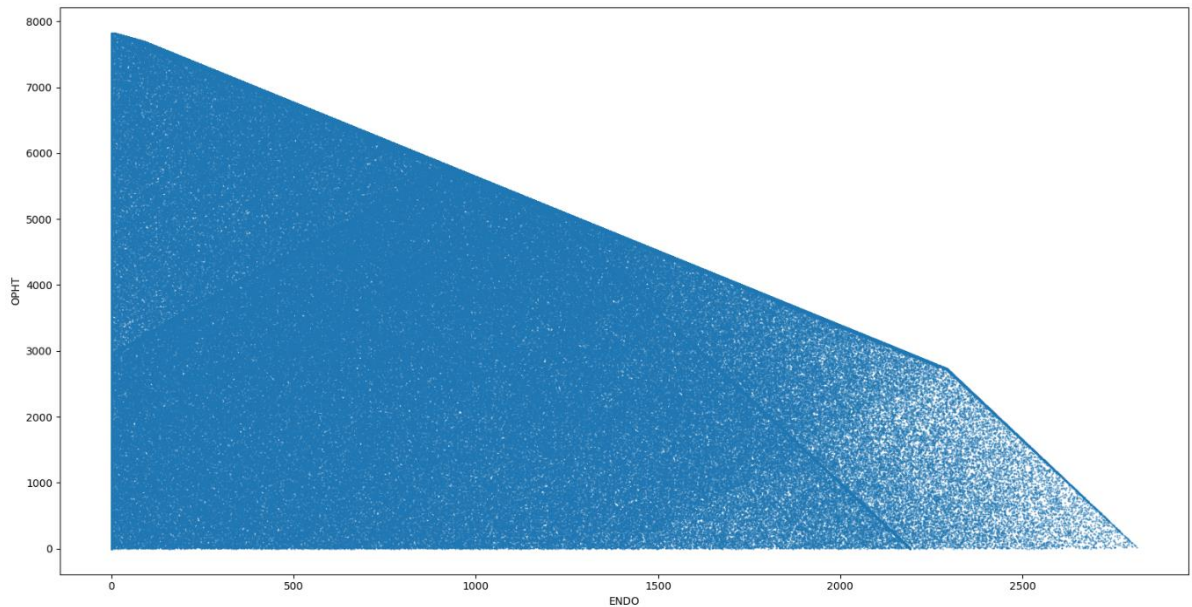

Cardiology versus Respiratory. These specialties share ward 5A. There is no diagonal trade-off region. The space in the top left is, however, quite empty. There are distinct vertical lines on both the left and right sides, and a dense region midway on the right side.

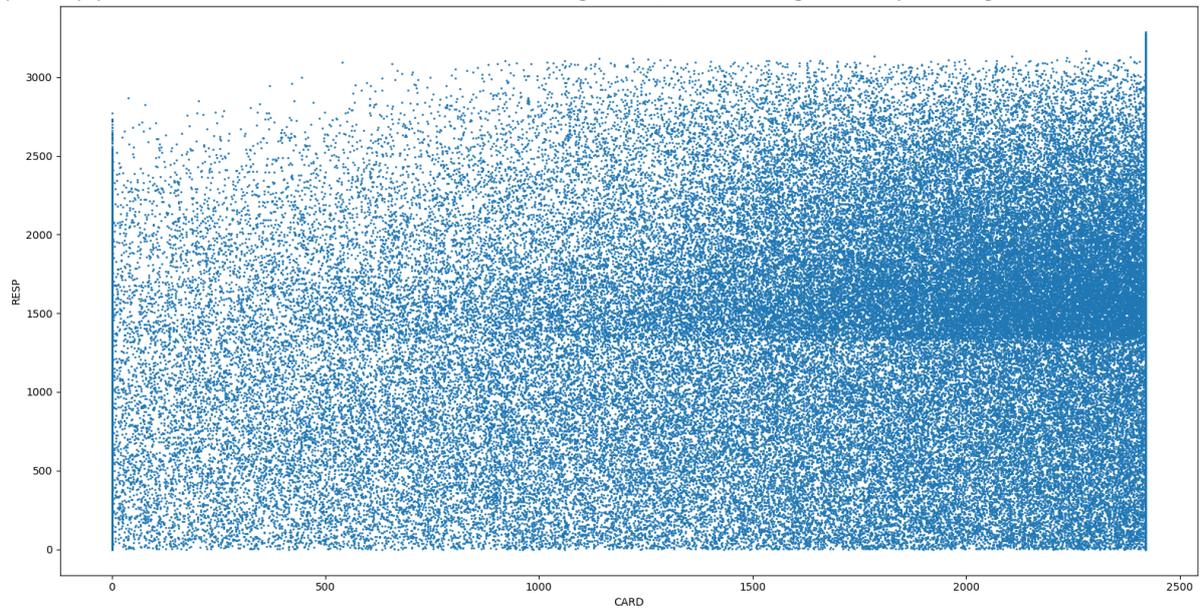

**Appendix F. Trade-offs Between Specialties (3D Projections)**

In this section, the 3-dimensional plots show the reason for some of the denser banded regions that were evident in Appendix E.

Endocrine versus Gastroenterology and Nepthrology. The landscape shown is very complex in this part of the objective space.

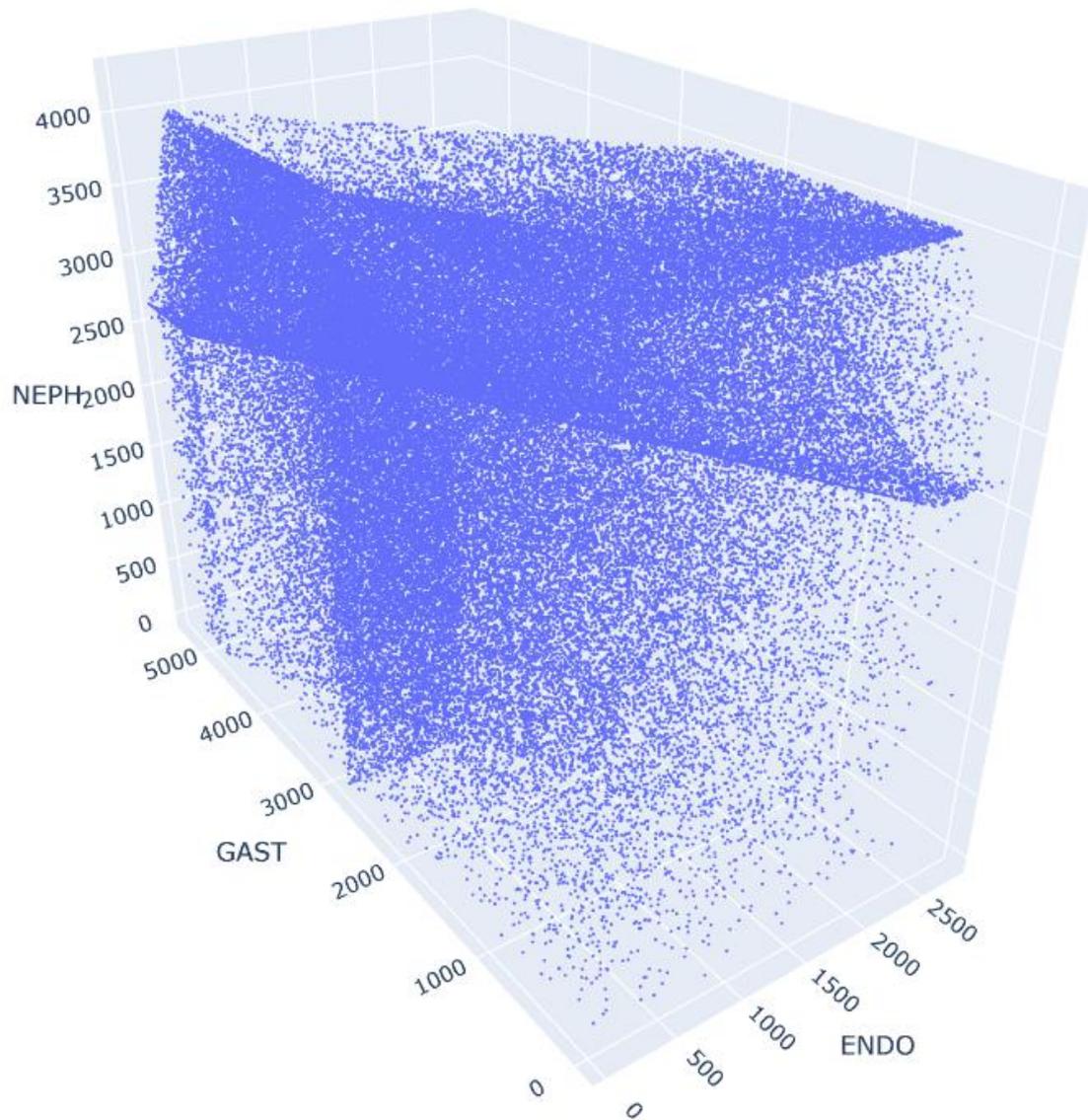

Gynaecology versus Gastroenterology and Hepatology. There are three distinct planes here.

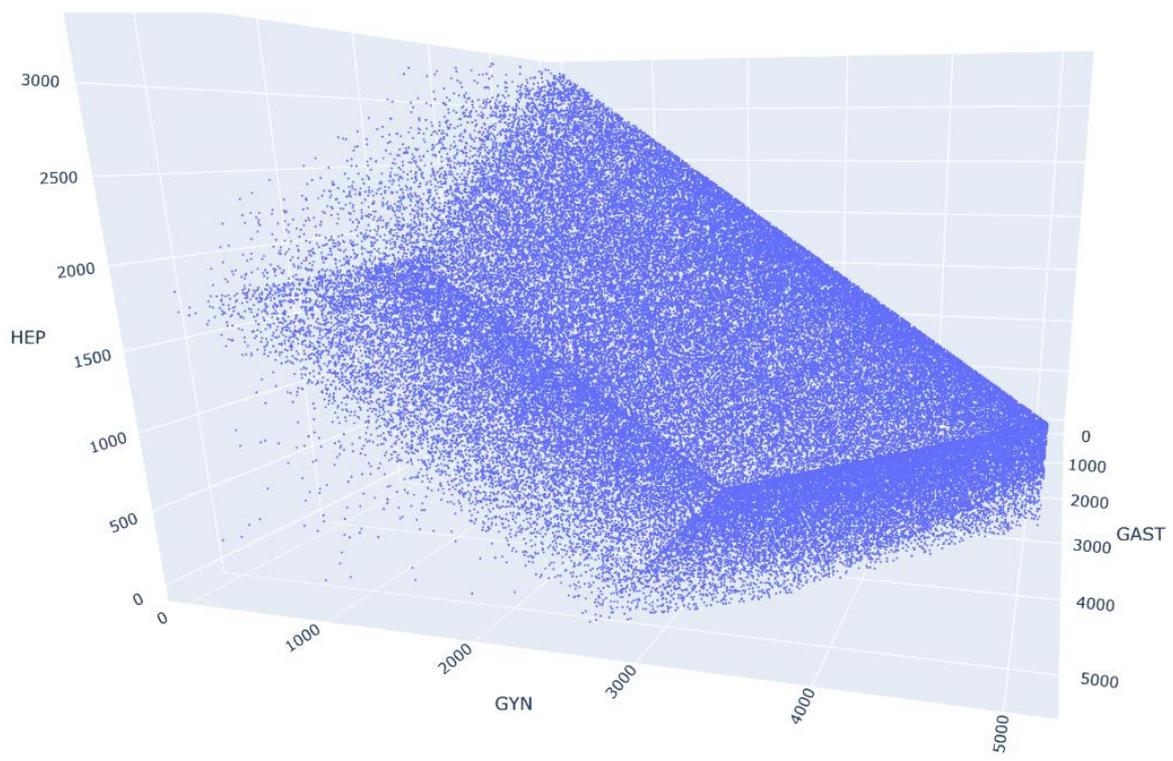

Ophthalmology versus Cardiology and Respiratory. There is a distinctive vertical plane with a greater density of solutions.

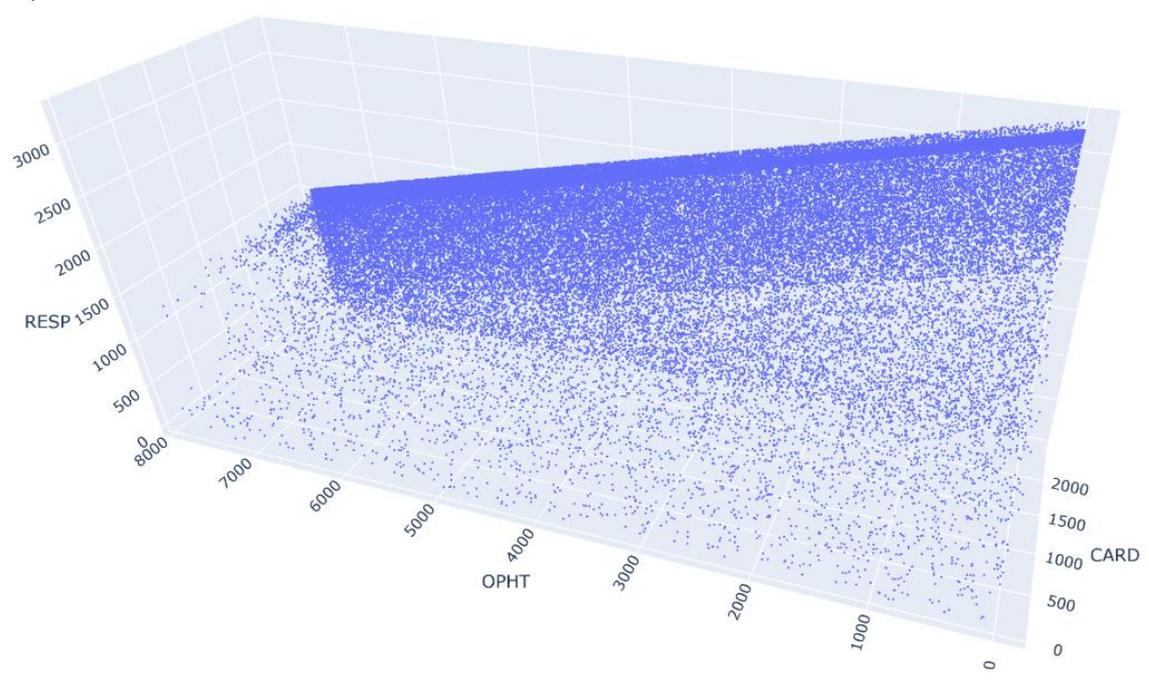

Respiratory versus Immunology and Neurology. There are two very distinctive planes intersecting here.

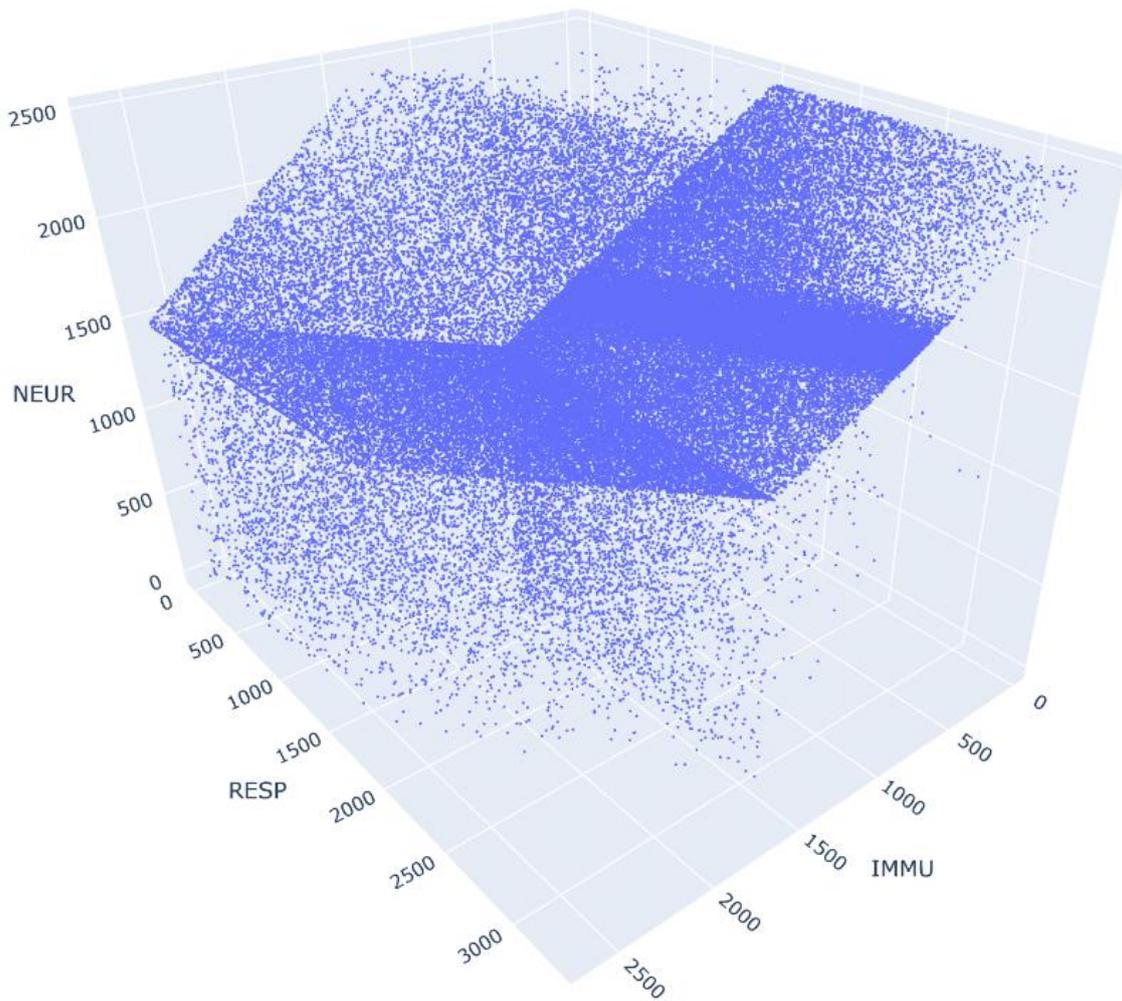

Gastroenterology versus Ophthalmology and Endocrine.

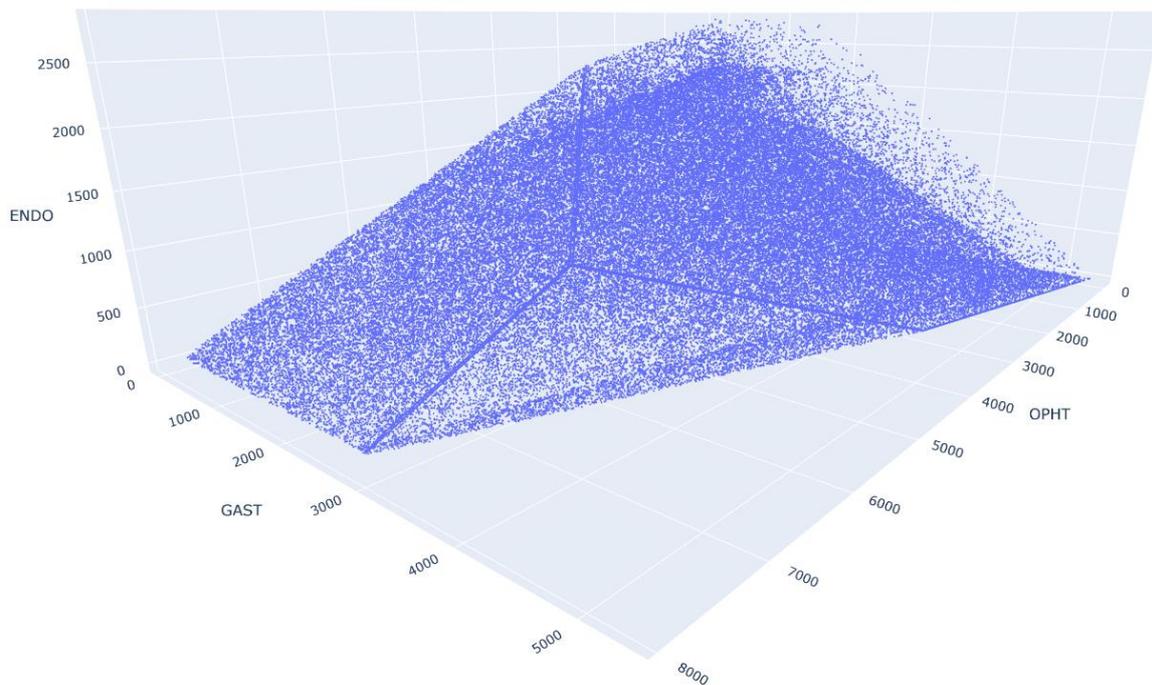